\documentclass[final,3p,times,twocolumn]{elsarticle}

\usepackage{amssymb}

\usepackage{preamble}

\journal{Arxiv}

\begin{document}

\begin{frontmatter}



\title{Hierarchical Multi-robot Strategies Synthesis and Optimization under Individual and Collaborative Temporal Logic Specifications }


\author{Ruofei Bai\fnref{fn1}}
\author{Ronghao Zheng$^{1, 2}$\corref{cor1}}
\ead{rzheng@zju.edu.cn}
\cortext[cor1]{Corresponding author.}
\author{Yang Xu\fnref{fn1}}
\author{Meiqin Liu$^{2, 3}$}
\author{Senlin Zhang$^{1, 2}$}

\fntext[fn1]{College of Electrical Engineering, Zhejiang University, Hangzhou, 310027, China}
\fntext[fn2]{State Key Laboratory of Industrial Control Technology, Zhejiang University, Hangzhou, 310027, China}
\fntext[fn3]{School of Electronics and Information Engineering, Xian Jiaotong University, Xian, 710049, China}


            
            

\begin{abstract}
This paper presents a hierarchical framework to solve the multi-robot temporal task planning problem.
We assume that each robot has its individual task specification and the robots have to jointly satisfy a global collaborative task specification, both described in linear temporal logic.
Specifically, a central server firstly extracts and decomposes a collaborative task sequence from the automaton corresponding to the collaborative task specification, and allocates the subtasks in the sequence to robots.
The robots can then synthesize their initial execution strategies based on locally constructed product automatons, combining the assigned collaborative tasks and their individual task specifications. 
Furthermore, we propose a distributed execution strategy adjusting mechanism to iteratively improve the time efficiency, by reducing wait time in collaborations caused by potential synchronization constraints. We prove the completeness of the proposed framework under assumptions, and analyze its time complexity and optimality. Extensive simulation results verify the scalability and optimization efficiency of the proposed method.
\end{abstract}




\end{frontmatter}


\section{Introduction}

Multi-robot task planning widely exists in many areas, such as smart logistics, autonomous inspection, intelligent manufacturing, etc. 
It remains a challenge to efficiently formalize and solve the task planning problems under complex task requirements.
Formal methods based on model checking theories, such as linear temporal logic (LTL), have draw increasing attention in recent years, due to its user-friendly syntax and expressive power in describing temporally constrained task specifications.
The temporal logic specification can be automatically transformed to correct-by-construction controller for robots, providing feedback and guarantees for robot behaviors~\cite{kress-2018-annunal_review}.
Given a global LTL task specification, the execution strategies for robots can be synthesized by searching on a constructed automaton that combines all robots' environment models and the automaton corresponding to the LTL formula.

In this paper, we focus on a situation where each robot has its individual finite LTL task specification, and the robots have to jointly satisfy a global collaborative finite LTL task specification. 
Each robot's individual tasks can be satisfied by itself, while the completion of collaborative tasks may require several robots of different types.
Prior studies of multi-robot task planning under locally given task specifications rely on either online reactive planning or centralized computation.
Most online methods assume that the collaboration requirements are integrated into the local LTL specification, so that the assignments of collaborative tasks are completely or partially known in prior.
In addition, the online methods may fail to optimize the performance from a global perspective.
The methods based on centralized computation can be applied to more general situations, but usually afford exponential complexity and thus has poor scalability.
While most existing temporal logic planning methods focus on the feasibility and the optimization of traveling distance, little attention has been paid to reduce the total time cost for robots.
This problem is particularly prominent in temporal logic tasks with potential synchronization constraints, because robots may spend extra wait time in collaborations and thus become inefficient.
The optimization of time cost considering wait time in collaborations has not been explicitly evaluated in previous works.

To mitigate the above issues, we propose a hierarchical multi-robot temporal task planning framework to synthesize feasible and optimized task execution strategies for robots.
We do not assume the collaborative tasks are allocated in prior, and try to minimize the total time cost for robots.

First, a central server extracts a task sequence which satisfies the collaborative task specification from the related automaton, and allocates the subtasks in the sequence to robots by solving a constructed SMT (Satisfiability Modulo Theories) model.
The SMT formulation includes collaboration requirements and synchronization constraints of tasks, and also enables the robots to overcome unstable communication links between some task regions in real deployment.
All feasible task assignments are iteratively generated and evaluated to continuously optimize the results, which can be easily implemented by the SMT formulation.
Then, given the assigned collaborative tasks, the robots modify their individual task specifications by integrating the assigned collaborative tasks as well as the related temporal constraints to them, and then locally constructs its product automaton based on model checking theories.
The initial execution strategy can then be obtained by searching the shortest on the automaton, which is optimal from individual robot's perspective.
However, the locally obtained initial execution strategies may be time inefficient when collaborating with other robots because of their different arrival time to collaborative tasks.
Finally, to optimize the total time cost, we propose a distributed execution strategy adjusting mechanism to iteratively improve the initial execution strategies based on inter-robot communication, which prevents the exponential complexity encountered in traditional methods based on product automaton.  Specifically, a token is passed around the robots, and the robot who get the token has the authority to adjust its execution strategy to reduce the time cost. The adjustment is motivated by sequentially traversing all collaborative tasks and greedily reducing wait time in each collaboration.
After a valid adjustment, the robot propagates its modified timeline to other robots, which includes its modified arrival time instances to the assigned collaborative tasks.

In our previous work~\cite{individual-and-collaborative_arxiv_2021}, we have proposed a preliminary framework to solve the problem.
We extend the work in~\cite{individual-and-collaborative_arxiv_2021} by introducing pruned local product automatons for robots and operating the execution strategy adjusting mechanism on them, which greatly improves the solving efficiency. 
Moreover, we design the distributed version of the proposed adjusting mechanism based on message exchanging, to distribute the computational burden among robots.
We also formalize the optimization of execution strategies as a mixed integer linear programming (MILP) to compute the optimally execution strategies for robots given the assignments of collaborative tasks. This is used as a baseline method and further verifies the high efficiency and solution quality of the proposed adjusting mechanism.
Finally, we prove the completeness of the proposed method under some assumptions.

The contributions of the paper are summarized as follows: 
\begin{enumerate}
 \item we propose a hierarchical multi-robot temporal task planning framework, which efficiently synthesizes execution strategies for robots satisfying both individual and global collaborative LTL specifications, without assuming that the collaborative tasks are pre-assigned to robots;
 \item we propose a distributed execution strategy adjusting mechanism, in which the robots can iteratively improve the performance of their execution strategies via inter-agent communication, reducing the inefficient wait time in collaborations caused by potential synchronization constraints;
 \item we prove the completeness of the proposed method under assumptions and analyze its time complexity. Extensive simulation experiments verify the scalability and efficiency of the proposed method.
\end{enumerate}   

\subsection{Related Works}

Existing studies in multi-robot task planning under temporal logic constraints fall into two categories: the top-down and the bottom-up patterns.
On the one hand, studies in top-down pattern usually assume a globally given temporal logic specification for a team of robots.
The task execution strategies of the robots can be obtained by searching for a path on a constructed product automaton that combines all robots' environment models and an automaton corresponding to the LTL formula. The construction of product automaton suffers from the exponential complexity. Also, the transformation from the LTL formula to the corresponding automaton has the exponential complexity w.r.t the length of the LTL formula.
Several approaches have been proposed to improve the scalability of the traditional method, such as pruning redundant states of the environment models and the product automaton~\cite{T*-A-Heuristic-ICRA-2020, kantarosIntermittentConnectivityControl2015};
 sampling-based construction and searching approaches of the product automaton~\cite{vasileSamplingBasedTemporalLogic2013, kantarosSamplingbasedControlSynthesis2017,kantarosDistributedOptimalControl2018, stylus-IJRR-2020}; and decomposition of the globally given task specification~\cite{schillingerDecompositionFiniteLTL2018, schillingerSimultaneousTaskAllocation2018, Faruq-2018-simultaneous_uncertainty, Banks-2020-cross_entropy}.
 Some abstraction-free temporal logic planning methods based on sampling strategy~\cite{ an-abstraction-TRO-2021} or reinforcement learning~\cite{schillinger2019hierarchical} have also been investigated.
 Besides, some works convert the temporal logic planning problem into a mixed integer linear programming model, and then the off-the-shelf optimizer like Gurobi can be utilized to solve it~\cite{7945014, multirobot-counting-TRO-2020}.
 
 On the other hand, studies in bottom-up pattern typically distribute the task specification to individual robots, and the robots jointly satisfy some global task requirements~\cite{decentralized-swarm-MRS-2017}. 
 M.~Guo~$et\ al.$~\cite{guoBottomupMotionTask2015} investigated the task coordination of loosely coupled multi-agent systems with dependent local tasks.
 The robots independently synthesize their off-line initial plans first, and then the collaborative actions in each robot's local tasks are performed with other robots' assistance, through online communication and computation. 
 The above method is further modified in \cite{guoTaskMotionCoordination2017} to include heterogeneous capabilities of robots and online task swapping mechanism. 
 J.~Tumova~$et\ al.$ \cite{tumovaDecompositionMultiagentPlanning2015} considered a slightly different setting from~\cite{guoBottomupMotionTask2015}, in which each robot have its local complex temporal logic specifications, including an independent motion specification and a dependent collaborative task specification.
 An two-phase automata-based method was proposed, where each robot's motion planning is synthesized based on a local automaton in a decentralized way.
 The local automaton is further pruned by removing states not related to the collaborative tasks.
 The method then constructs a centralized product automaton of the pruned local automaton, on which the collaboration strategies can be obtained.
 Despite that the sizes of local automatons have been greatly reduced by the pruning techniques, the method still has exponential complexity w.r.t. the robot number.
Y. Kantaros $et\ al$.~\cite{kantarosTemporalLogicTask2019} analyzed a similar situation considered in this paper, that individual robots have their independent individual task specifications, while they have to jointly satisfy a globally given collaborative task specification.
Despite that, the collaborative task specification in~\cite{kantarosTemporalLogicTask2019} is restricted to intermittent communication tasks, that require several robots to intermittently gather at some locations to exchange information.
The gathering locations are iteratively modified by an on-line optimization procedure to reduce the total travel distance.
A common assumption of the aforementioned studies in bottom-up pattern is that the allocation of collaborative tasks are fully or partially known in prior.

Inspired by~\cite{kantarosTemporalLogicTask2019}, we extend the global collaborative task specification into finite LTL; and aim to minimize the total time spent to finish all tasks considering inefficient wait time in collaborations, which has not been considered in~\cite{kantarosTemporalLogicTask2019}.
We do not assume the allocation of collaborative tasks are explicitly given, and formulate an SMT model to flexibly address the task allocation problems and possible communication limitations in real execution.

The rest of this paper is organized as follows. In Sec.~\ref{sec: preliminaries}, we introduce the linear temporal logic, task specification, and formalize the problem we considered.
Then in Sec.~\ref{sec: framework}, we propose the hierarchical multi-robot temporal task planning framework to compute execution strategies for robots.
An execution strategy adjusting mechanism is proposed in Sec.~\ref{sec: optimization} to further improve the time efficiency of execution strategies for robots.
The results of experiments are analysed in Sec.~\ref{sec: experiment}. Finally, we draw our conclusion in Sec.~\ref{sec: conclusion}.

\section{Preliminaries and Problem Formulation}
\label{sec: preliminaries}

\subsection{Linear Temporal Logic}

An LTL formula $\varphi$ over a set of atom propositions $\mathcal{AP}$ are defined according to the following recursive grammar~\cite{baierPrinciplesModelChecking2008}:
\[\varphi::={\rm true}\ |\ \pi\ |\ \varphi_{1}\wedge \varphi_{2}\ |\ \neg \varphi\ |\ \circ \varphi\ |\ \varphi_{1} \rm U \varphi_{2},\] 
where $\mathrm{true}$ is a predicate true and $\pi\in \mathcal{AP}$ is an atom proposition. 
The other two frequently used temporal operators $\Diamond$~(eventually) and $\Box$~(always) can be derived from operator ${\rm U}$~(until). 
In this paper, we only consider a kind of finite LTL~\cite{giacomoReasoningLTLFinite} called ${\rm LTL}_{f}$, which is a variant of original LTL that can be interpreted over finite sequences, and uses the same syntax as original LTL.
Moreover, we exclude the $\bigcirc$ (next) operator from the syntax, since it is not meaningful in practical applications~\cite{kloetzerFullyAutomatedFramework2008}.
We refer the reader to \cite{baierPrinciplesModelChecking2008} for the details of LTL semantics.

Given an ${\rm LTL}_{f}$ formula $\varphi$, a nondeterministic finite automaton (NFA) can be constructed which accepts exactly the sequences that make $\varphi$ true.

\begin{definition}[Nondeterministic Finite Automaton]
	A nondeterministic finite automaton (NFA) $\mathcal{F}$ is a tuple $\mathcal{F} := \langle \mathcal{Q}_{F},\mathcal{Q}_{F}^{0}, \alpha, \delta, \mathcal{Q}_{F}^{F}\rangle$, where 
	$\mathcal{Q}_{F}$ is a finite set of states;
	$\mathcal{Q}_{F}^{0}\subseteq \mathcal{Q}_{F}$ is a set of initial states;
	$\alpha$ is a set of Boolean formulas over $\pi \in \AP$;
	$\delta:\mathcal{Q}_{F}\times \mathcal{Q}_{F}\rightarrow \alpha$ is the transition condition of states in $\mathcal{Q}_{F}$; 
	$\mathcal{Q}_{F}^{F}$ is a set of accepting states.
\end{definition}

A finite sequence of states $q_{F}\in \mathcal{Q}_{F}$ is called a run $\rho_{F}$. A run $\rho_F$ is called \emph{accepting} if it starts from one initial state $q_{F}^{0}\in \mathcal{Q}_{F}^{0}$ and ends at an accepting state $q_{F}^{F}\in \mathcal{Q}_{F}^{F}$. 
Given a finite sequence $\sigma = \sigma(1)\sigma(2)\dots\sigma(L-1)\sigma(L)$, we say that $\sigma$ describes a run $\rho_F$ if $\sigma(i)\vDash \delta\left(\rho_{F}(i), \rho_{F}(i+1)\right)$ for all $r\in \left[1:L-1\right]$, i.e., $\sigma(i)$ is a set of atom propositions or negative propositions, which makes the Boolean formula $\delta\left(\rho_{F}(i), \rho_{F}(i+1)\right)$ become ${\rm true}$.
Here $\left[1:L-1\right]$ denotes a set of indexes increasing from $1$ to $L-1$ by step $1$.
A finite sequence $\sigma$ fulfills an ${\rm LTL}_{f}$ formula if at least one of its runs is accepting.
The minimum requirements for a sequence to describe a run can be represented as an essential sequence, as in the following definition.
\begin{definition}[Essential Sequence]
\label{essential-sequence}
Considering a sequence $\sigma = \sigma(1)\dots\sigma(L)$, it is called \emph{essential} if and only if it describes a run $\rho_F$ in NFA $\mathcal{F}$ and $\sigma(i)\backslash \{\pi\} 
\nvDash \delta(\rho_{F}(i), \rho_{F}(i+1))$ for all $r\in \left[1:L-1\right]$ and $\pi\in \mathcal{AP}$.
\end{definition}

\subsection{Task Requirements}
\label{ts_ltl}

\begin{figure}[t]
	\flushleft
	\begin{overpic}
		[scale=0.125]{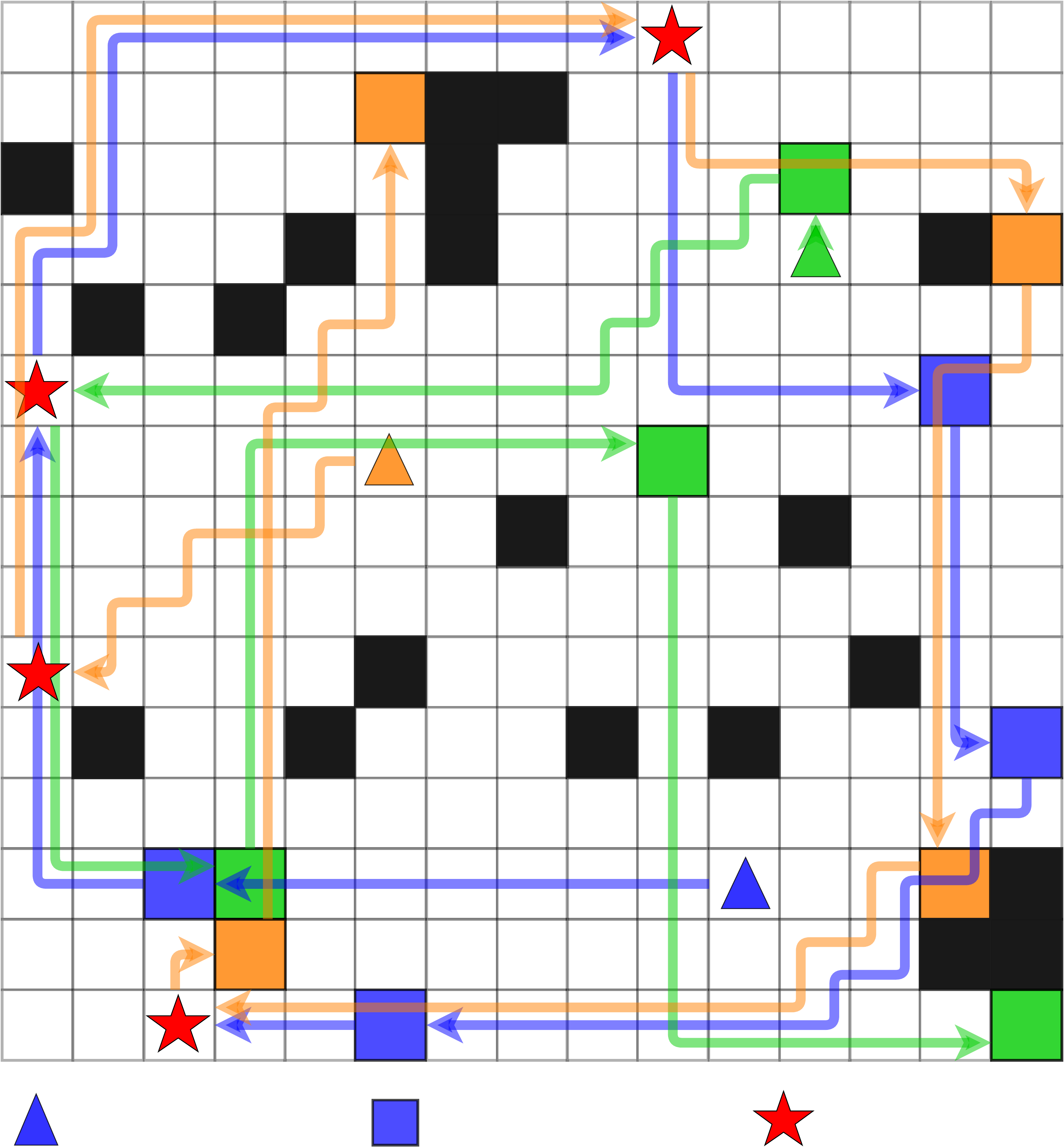}
        \put(100, 85){\scriptsize $\varphi_{1} = (\Diamond \pi^{ts_{1}}) \wedge (\Diamond \pi^{ts_{2}}) \wedge (\Diamond \pi^{ts_{3}}) \wedge$}
        \put(117, 75){\scriptsize $ (\Diamond \pi^{ts_{4}}) \wedge (\neg \pi^{ts_{1}} ~{\rm U}~ \pi^{ts_{4}})$}
        \put(100, 65){\scriptsize $\varphi_{2} = \cdots$}
        \put(100, 55){\scriptsize $\dots$}
        \put(100, 25){\scriptsize $\phi = (\Diamond \pi^{ct_{1}}) \wedge (\Diamond \pi^{ct_{2}}) \wedge (\Diamond \pi^{ct_{4}}) \wedge $}
        \put(113,15){\scriptsize $(\neg \pi^{ct_{3}}~{\rm U}~\pi^{ct_{2}})  \wedge$}
        \put(113,5){\scriptsize $(\Box(\pi^{ct_{4}} \rightarrow (\Diamond \pi^{ct_{3}})))$}
        \put(8,0){\scriptsize robot $r$}
        \put(40,0){\scriptsize $ts\in \overline{\mathbb{T}}_i$}
        \put(74,0){\scriptsize $ct\in \widetilde{\mathbb{T}}$}
	\end{overpic}
	\caption{The simulation results of am experiment with $3$ robots.}
	\label{environment}
\end{figure}
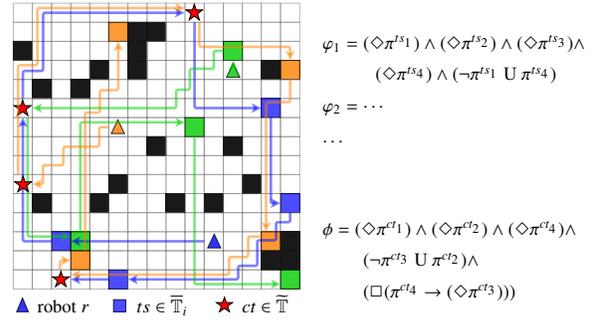

Consider there is a set of robots $\mathcal{N}:=\{1, ..., N\}$ operating in an environment, e.g., as in Fig.~\ref{environment}, which can be represented as a graph $\mathcal{W} = \langle\mathcal{Q}, \mathcal{E}\rangle$.
Here $\mathcal{Q}$ is the set of regions of interest and $\mathcal{E}$ contains the connectivity relations of regions in $\mathcal{Q}$. 
Each robot $r\in \mathcal{N}$ has its specific capability $c_j\in Cap$ and thus can perform specific actions in the environment $\mathcal{W}$. 
The set $Cap:=\{c_j\}_{j\in \{1, ..., |Cap|\}}$ contains all capability types.
We define $\N_j:=\left\{r~|\text{robot $r$ has the capability $c_j$}\right\}$.

There are several tasks to be completed by the robots, which are distributed in $\W$, as defined in Definition~\ref{def-1}.

\begin{figure*}[t]
   \centering
    \includegraphics[width=\linewidth]{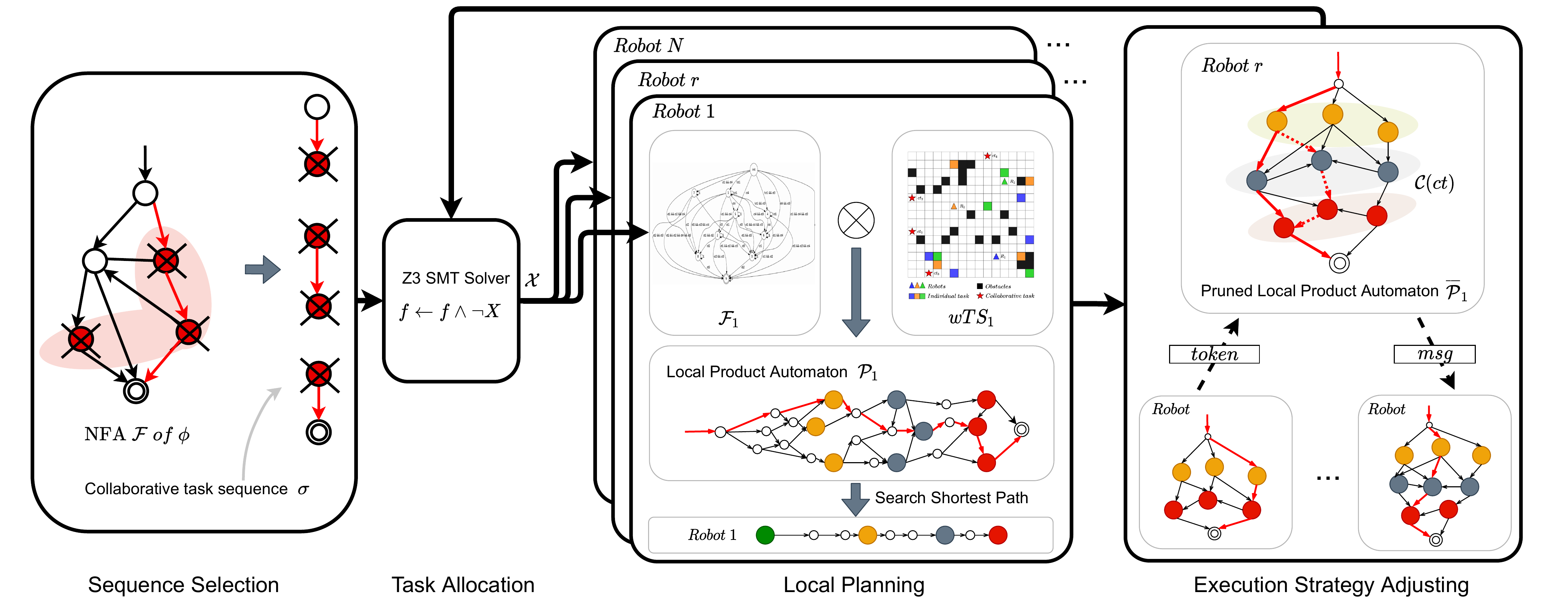}
    \caption{The proposed framework for multi-robot temporal logic task planning.}
    \label{fig:framework}
\end{figure*} 

\begin{definition}[Task Requirement]
\label{def-1}
A task in $\mathcal{W}$ is a tuple $ts := \langle \pi^{ts}, \{(c_j, m_j^{ts})\}_{j\in I_{ts}}, q^{ts}\rangle$, where $\pi^{ts}$ is the unique atom proposition of $ts$; $q^{ts}\in \mathcal{Q}$ corresponds to the region associated with $ts$; and $ts$ can be completed, i.e, $\pi^{ts}$ becomes $\mathrm{true}$, if at least $m_{j}^{ts}$ robots with capability $c_{j}$ are deployed simultaneously in the region $q^{ts}$ and perform specific actions according to their capabilities, for all $j\in I_{ts}$.
Here $I_{ts}\subseteq \{1, ..., |Cap|\}$ collects indexes of the capabilities required by $ts$.
\end{definition}

In this paper, we consider the situation that: (1) each robot $r\in \mathcal{N}$ has its pre-assigned individual task specifications $\varphi_{r}$ that can be satisfied by itself. 
Here $\varphi_{r}$ is an ${\rm LTL}_{f}$ formula defined over
$\overline{\mathbb{T}}_r$, which is robot $r$'s individual task set.
For each $ts\in \overline{\mathbb{T}}_r$, it holds that $|I_{ts}|=1$, $m^{ts}_j=1$, and $r\in \mathcal{N}_j$;
(2) the robots have to jointly satisfy a global ${\rm LTL}_{f}$ task specification $\phi$ defined over a set of tasks $\widetilde{\mathbb{T}}$, in which each task $ct\in \widetilde{\mathbb{T}}$ may has much heavier workloads and requires no less than one robot with several different capabilities.
Note that we particularly use $ct$ to represent a task in $\widetilde{\mathbb{T}}$, called \emph{collaborative task}, to distinguish it from tasks in the set $\bigcup_{r\in \mathcal{N}}\overline{\mathbb{T}}_r$ in the following context.

We assume that the executions of individual task specification $\varphi_{r}$ are independent of each other, i.e., do not influence the states of other robots. 
And they are also independent of the execution of collaborative task specification $\phi$.
More formally, we assume that: $\forall~i, j\in \mathcal{N}$ and $i\ne j$, $\overline{\mathbb{T}}_i\cap \overline{\mathbb{T}}_j=\emptyset$; and $\forall~r\in\mathcal{N}$, $\widetilde{\mathbb{T}}\cap \overline{\mathbb{T}}_r=\emptyset$.
The assumption is reasonable because if some individual tasks and collaborative tasks are non-independent, these task requirements can be formulated into the global collaborative task specification $\phi$. 

The mobility and capability of each robot $r$ can be formalized as a weighted transition system.

\begin{definition}[Weighted Transition System]
A weighted transition system $wTS$ of robot $r$ is a tuple $wTS_{r} = \langle \mathcal{Q}_r, q_{r}^{0}, \\ \rightarrow_{r}, \omega_{r}, \mathcal{AP}_{r}, L_{r} \rangle$, where $\mathcal{Q}_{r}\subseteq \mathcal{Q}$ is a finite set of states corresponding to regions in $\mathcal{W}$; $q_r^0$ is the initial state; $\rightarrow_{r}\subseteq \mathcal{E}$ contains all pairs of adjacent states; $\omega_{r}: \mathcal{Q}_{r}\times \mathcal{Q}_{r} \rightarrow \mathbb{R}^{+}$ is a weight function measuring the cost for each transition; $\mathcal{AP}_{r}$ is the set of atom propositions related to the tasks in $\mathcal{W}$; $L_{r}: \mathcal{Q}_{r} \rightarrow 2^{\mathcal{AP}_{r}}$ is a labeling function, 
and satisfies that $(1)$ $\forall~ts\in \overline{\mathbb{T}}_r$, $\pi^{ts}\in L_{r}(q^{ts})$; $(2)$ $\forall~ct\in \widetilde{\mathbb{T}}$, $\pi^{ct}\in L_{r}(q^{ct})$ iff $\exists j\in I_{ct}$, $r\in \mathcal{N}_j$.
\end{definition}

A task execution strategy $\tau_r$ for each robot $r$ is a \emph{walk} in the $wTS_{r}$. The execution of $\tau_r$ will induce a sequence $\sigma_i$ of sets of atom propositions, which follows the labeling function $L_{r}$.

\subsection{Problem Formulation}

Given task execution strategy $\tau_r$ for each robot $r$, 
the robots execute their plans asynchronously and synchronize with other assistant robots when performing the collaborative tasks.
The total time cost to satisfy all task requirements is defined as 
$
T^{\rm colla} = \sum_{r\in \mathcal{N}} T^{\rm colla}_{r},
$
where $T^{\rm colla}_{r}$ denotes the amount of time robot $r$ spends to execute $\tau_r$, considering the wait time in each collaboration due to potential synchronization constraints. 
The problem to minimize the total time can be formalized as: 

\begin{problem}
\label{problem1}
Consider a set $\mathcal{N}$ of robots operate in an environment  $\mathcal{W}$, and suppose that the mobility and capability of each robot $r$ is modeled as a weighted transition system $wTS_{r}$. 
Given individual ${\rm LTL}_{f}$ task specification $\varphi_{r}$ for each robot $r$ and a global collaborative task specification $\phi$, find task execution strategy $\tau_{r}$ in $wTS_{r}$ for each robot $r$ which satisfies:
\begin{enumerate}
    \item  the execution of $\tau_{r}$ satisfies individual task specifications $\varphi_{r}$;
    \item the joint behaviors of robots $r$ satisfies collaborative task specification $\phi$; 
    \item the total time cost $T^{\rm colla}$ is minimized.
\end{enumerate}
\end{problem}

\section{Multi-Robot Temporal Task planning}
\label{sec: framework}

 In this section, we propose the multi-robot temporal task planning framework as shown in Fig.~\ref{fig:framework} to solve Problem~\ref{problem1}. 
 The proposed framework firstly finds a sequence of collaborative tasks which satisfies the collaborative task specification and allocates the tasks in the sequence to robots.  
 Then each robot locally synthesizes its initial task execution strategy which satisfies the individual task specification and the assigned collaborative tasks requirements. 
 Finally, a distributed execution strategy adjusting mechanism is proposed in Sec.~\ref{sec: optimization} to optimize the initial task execution strategies. 
 All feasible task assignments are iteratively generated and evaluated to continuously optimize the results.
 The overall multi-robot temporal task planning framework is shown in Alg.~\ref{alg-framework}.

 \subsection{Selection and Decomposition of Execution Sequence}
 \label{3A}
 
 To find a collaborative task sequence satisfying the collaborative task specification $\phi$, a three-step
 scheme is used:

  \begin{enumerate}
      \item prune the NFA $\mathcal{F}$ of $\phi$ by removing all impossible transitions that require assistance beyond what the robot set $\mathcal{N}$ can provide.
      \item identify the decomposition states in the pruned $\mathcal{F}$ by utilizing the decomposition algorithm proposed in~\cite{schillingerDecompositionFiniteLTL2018}.
      \item select an essential sequence which describes an accepting run in $\mathcal{F}$, and divide it into independent subsequences by the decomposition states along the run. 
  \end{enumerate}

The way to prune $\mathcal{F}$ is straightforward by checking the atom propositions required to be true in each transition condition and the details are omitted here. 

The essential sequence $\sigma$ that describes an accepting run $\rho_{F}$ in the pruned NFA $\mathcal{F}$ is actually a feasible collaborative task sequence that satisfies $\phi$.
Given an accepting run $\rho_{F}$ of $\mathcal{F}$ and its corresponding essential sequence $\sigma$, the \emph{decomposition states} along the run $\rho_{F}$ can decompose $\sigma$ into $S$ $(\ge 1)$ subsequences, i.e., $\sigma = \sigma^{1} ;\sigma^{2};\dots;\sigma^{S}$.
Let $\mathcal{S}:=\left[1:S\right]$ denote the set of indexes.
Such decomposition is proved to have two properties in \cite{schillingerSimultaneousTaskAllocation2018}: (1) \textbf{Independence}. Execution of each subsequence $\sigma^{k}$ will not violate another $\sigma^{j}$, $\forall~k, j\in \mathcal{S}$ and $k\ne j$; (2) \textbf{Completeness}. The completion of all subsequences $\sigma^{k}$ implies the completion of $\sigma$.
The detailed definition of decomposition states can be found in Definition~$9$ and Theorem $2$ of \cite{schillingerDecompositionFiniteLTL2018}.

The $m$-th element $\sigma^{k}(m)$
in $\sigma^{k}$ is defined as $\sigma^{k}(m) := \{ct^{k}_{(m, j)}\}_{j\in \left[1:|\sigma^{k}(m)|\right]}$, in which $ct^{k}_{(m,j)}\in \widetilde{\mathbb{T}}$ denotes the $j$-th task in the $m$-th element of $\sigma^{k}$.
For notation simplicity, we use $ct$ and $\pi^{ct}$ interchangeably afterward.
Note that we only consider the positive atom propositions in the essential sequence $\sigma$, and assume that the negative propositions are checked and guaranteed in online execution.
Although it may impose extra implicit communications in real deployment, it can greatly simplify the task planning procedure.
Similar assumption can also be found in \cite{schillingerDecompositionFiniteLTL2018, luo-2021-temporal}.

The execution of each sub-sequence $\sigma^{k}$ is independent, while the execution of collaborative tasks within each $\sigma^{k}$ must satisfy the following temporal constraints according to the accepting condition of $\mathcal{F}$.
\begin{itemize}
    \item  \textbf{Synchronization Constraints}. ${\forall}\ ct^{k}_{(m, j)}$, 
    $ct^{k}_{(m, j')}\in \sigma^{k}(m)$, $j\ne j'$, 
    $ct^{k}_{(m, j)}$ and $ct^{k}_{(m, j')}$ need to be executed synchronously;
    \item   \textbf{Ordering Constraints}. $\forall~ct^{k}_{(m, j)}$, $ct^{k}_{(m', j')}\in \sigma^{k}, m<m'$, 
    $ct^{k}_{(m', j')}$ must be executed after $ct^{k}_{(m, j)}$.
\end{itemize}

The above decomposition of essential sequence relaxes the temporal constraints between the tasks in the essential sequence, which may help to reduce the unnecessary communication as well as the total time cost in real deployment, as stated in Remark~\ref{remark-decomposition} of Sec.~\ref{3c}.

The selection of the collaborative task sequence $\sigma$ has implicit influences on the performance of the final results.
It is time-consuming to investigate all feasible $\sigma$. 
Intuitively, the essential sequence related to an accepting run $\rho_{F}$ in $\mathcal{F}$ which has less coupling in each transition, contains as many decomposition states as possible, and has the smallest possible length, may help to improve the execution performance.  
To reduce the complexity, in the sequel we just select the essential sequence $\sigma$ that describes one of the shortest accepting run $\rho_{F}$ in the pruned $\mathcal{F}$ to be the collaborative task sequence, and decompose it by the decomposition states along the run $\rho_{F}$. 
The simulation results illustrate that the proposed method can find solutions of high quality in practice under the selected $\sigma$.

\subsection{SMT-Based Collaborative Task Allocation}
\label{subsec: smt}

After obtaining the decomposed collaborative task sequence $\sigma = \sigma^{1} ;\sigma^{2};\dots;\sigma^{S}$ and the related accepting run $\rho_{F}$ of $\mathcal{F}$, we now consider the allocation of collaborative tasks in the sequence $\sigma$ to satisfy the amount and types of assistance needed to complete each task, while following the temporal constraints implied by $\phi$.

For notation simplicity, we give all $ct^{k}_{(m,j)}\in \sigma^{k}$ an index $l $, defined as $ct^{k}_{l}:=ct^{k}_{(m, j)}$, where $l = \sum_{i = 1}^{m-1} |\sigma^{k}(i)| + j$, $\forall~m\in \left[1:|\sigma^{k}|\right]$ and $k\in \mathcal{S}$.
Then the $l$-th task in $\sigma^{k}$ can be referred as $ct^{k}_{l}$.
We define a set of Boolean variables $\mathcal{X}_{i} := \{
x_{r}^{(k, l)}|ct^{k}_{l}\in \sigma^{k}, k\in \mathcal{S}
\}$ for each robot $r$ to indicate the task assignment results. 
A true $x_{r}^{(k, l)}$ implies that robot $r$ is assigned to complete the task $ct^{k}_{l}$.

Now we construct the SMT model of the task allocation problem. 

(1) $\textbf{Collaboration Requirements.}$ A feasible task assignment must satisfy the amount and types of assistance needed to complete each collaborative task, as the following constraints:
For each $\sigma^k$,
${\forall}\ ct^{k}_{l}\in \sigma^{k}$,
\[
\sum_{r\in \mathcal{N}} \mathbf{1}\left(x_{r}^{(k, l)} \bigwedge r\in \mathcal{N}_{j}\right ) \ge m_{j}^{ct}, {\forall} j \in I_{ct} 
\]
is satisfiable, where $\mathbf{1}(\cdot)$ is an indicator function defined as $\mathbf{1}({\rm true}) = 1$ and $\mathbf{1}({\rm false}) = 0$.

(2) $\textbf{At Most One at a Time.}$ Each robot cannot participate in two distinct collaborative tasks which are executed synchronously. The constraints can be captured by: For each $\sigma^k$,
${\forall}\ ct^{k}_{l_{1}}, ct^{k}_{l_{2}}\in \sigma^{k}(m), l_{1}\ne l_{2}$,
\[
\neg x_{r}^{(k,l_{1})} \bigvee \neg x_{r}^{(k,l_{2})} , {\forall}\ r\in \mathcal{N} 
\]
is satisfiable.

(3) $\textbf{Communication Reduction.}$ The intersection of two sets of robots that complete two consecutive collaborative tasks is not empty: For each $\sigma^k$,
\[\bigvee_{r\in \mathcal{N}} \left(\vee_{l \in  \{l^{k}_{m-1}+1,..., l^{k}_{m}\}} x_{r}^{(k,l)}  \bigwedge \vee_{l \in  \{l^{k}_{m}+1,..., l^{k}_{m+1}\}} x_{r}^{(k,l)} \right)\]
is satisfiable, where $l^{k}_{m} = \sum_{j = 1}^{m}|\sigma^{k}(j)|$, $\forall~m\in [1:|\sigma^{k}|-1]$.

When the communication is limited in the environment, the above constraints $(3)$ can be applied to guarantee that each robot $r$ only needs to communicate with its assistant robots $j\in \mathcal{R}(ct)\backslash \{r\}$ when executing the assigned collaborative task $ct$.
The function $\mathcal{R}:ct^{k}_{l}\rightarrow 2^{\mathcal{N}}$ maps $ct^{k}_{l}$ to a set of robots that perform the task.
The constraint $(3)$ requires that $\mathcal{R}(\sigma^{k}(m))\bigcap \mathcal{R}(\sigma^{k}(m+1))\ne\emptyset$, $\forall~m\in \left[1:|\sigma^{k}|-1\right]$. 
The robot $r\in \mathcal{R}(\sigma^{k}(m))\bigcap \mathcal{R}(\sigma^{k}(m+1))$ behaves like a coordinator to guarantee the execution order of tasks in  $\sigma^{k}(m)$ and $\sigma^{k}(m+1)$.
Here the function $\mathcal{R}$ is reloaded as $\mathcal{R}(\sigma^{k}(m)) = \bigcup_{ct\in \sigma^{k}(m)} \mathcal{R}(ct)$ according to SMT constraint (2) and the synchronization constraints in Sec.~\ref{3A}.

\begin{remark}
    The SMT constraints $(3)$ can be selectively applied to some pairs of consecutive collaborative tasks, whose locations may be far apart so that it is difficult to ensure the connectivity of the communication links between these regions. 
\end{remark}

The overall SMT formula $f$ is a conjunction of all Boolean expressions of the above constraints $(1)\,(2)\,(3)$.
We define $X = \bigwedge_{r\in \mathcal{N}}\left(\wedge_{x_{r}^{(k,l)}\in \mathcal{X}_{i}}x_{r}^{(k,l)}\right)$ as the Boolean formula corresponds to a feasible task assignment $\{\mathcal{X}_r\}_{r\in \mathcal{N}}$ which satisfies~$f$. 
By utilizing an SMT solver, we can iterate all valid assignments by adding the negation of current $X$ into $f$, i.e., $f = f\wedge\neg X$. Therefore the solver will not return the same task assignment in the subsequent iterations, as shown in Line $15$ of  Alg.~\ref{alg-framework}.
Each feasible task assignment $\{\mathcal{X}_r\}_{r\in\mathcal{N}}$ will be passed into the subsequent local planning procedure to further investigate its performance.

Here we come up with a filtering strategy to fast filter the non-optimal task assignments. The efficiency of the filtering strategy will be shown in Sec.~\ref{sec: experiment}.

$\textbf{Filtering Strategy:}$ 
If ${\exists}\ m<n$ and $m, n\in \mathbb{N}^{+}$, such that ${\forall} r\in \mathcal{N}$, $\mathcal{X}^{n}_{r}\ge \mathcal{X}^{m}_{r}$, then $f$ can be directly modified as $f\wedge \neg X^{n}$ without proceeding the subsequent investigation procedures, where $\mathcal{X}^{m}_{r}$, $\mathcal{X}^{n}_{r}$ are the feasible task assignments
for robot $r$ in $m$-th and $n$-th iterations, respectively. 
Here $\mathcal{X}^{n}_{r}\ge \mathcal{X}^{m}_{r}$ means that $\forall~^n{x}_{r}^{(k,l)}\in \mathcal{X}^{n}_{r}$ and $^{m}{x}_{r}^{(k,l)}\in \mathcal{X}^{m}_{r}$,
$\mathbf{1}(^n{x}_{r}^{(k,l)})\ge \mathbf{1}(^{m}{x}_{r}^{(k,l)})$.

Given the obtained task assignment results, each robot will combine the assigned collaborative tasks with its local task specification to synthesize its initial execution strategy. 
The details will be discussed in the next subsection.

\subsection{Local Plan Synthesis With Collaborative Tasks}
\label{3c}
To synthesize the local task execution strategy, the key problem is how to construct a local ${\rm LTL}_f$ formula that captures the local individual task specification $\varphi_{r}$, as well as the temporal constraints of the assigned collaboration tasks.

Given one feasible task assignments $\mathcal{X}$, we construct set $\widetilde{\mathbb{T}}_{r}^{k} := \left\{ct^{k}_{l}\ \big\vert\ x_{r}^{(k, l)}~{\rm is~true}\right\}$ for all $k$ and $r$.
Let $\widetilde{\mathbb{T}}_{r} :=\bigcup_{k\in \mathcal{S}} \widetilde{\mathbb{T}}_{r}^{k}$. We sort all tasks in $\widetilde{\mathbb{T}}_{r}$ 
in the increasing order of indexes $k$ and $l$.

The modified local ${\rm LTL}_f$ formula $\widetilde{\varphi}_r$ for each robot $r$ is the conjunction of $\varphi_{r}$ and $\phi_{r}$, i.e., $\widetilde{\varphi}_r = \varphi_{r} \bigwedge \phi_{r}$. 
Here $\phi_{r}$ is the ${\rm LTL}_f$ formula that captures the requirements of tasks $ct\in \widetilde{\mathbb{T}}_{r}$.
There are two steps to construct $\phi_{r}$ and $\widetilde{\varphi}_r$:

\textbf{Step 1:} Formalize ordering constraints within each independent subsequence.  
We initialize the collaborative task specification $\phi_{r}^{k}$ corresponds to $\widetilde{\mathbb{T}}_{r}^{k}$ as $\phi_{r}^{k} = \Diamond(ct^{k}_{l_{1}})$, where $ct^{k}_{l_{1}}$ is the first task in the sorted $\widetilde{\mathbb{T}}_{r}^{k}$.
Then we use $ct^{k}_{l_{m}} \bigwedge \Diamond (ct^{k}_{l_{m+1}})$ to iteratively substitute $ct^{k}_{l_{m}}$ in $\phi_{r}^{k}$, where $ct^{k}_{l_{m}}$ and $ct^{k}_{l_{m+1}}$ are two consecutive tasks in sorted $\widetilde{\mathbb{T}}_{r}^{k}$, and $l_{m} < l_{m+1}$. 

The asynchronous execution of several independent subsequences may stick into deadlock in real deployment, due to the different execution order in each robot's local execution strategy. We use Step~$2$ to prevent the deadlock.

\textbf{Step 2:} Determine the execution order of each independent subsequence. For each robot $r$, we sort $\phi_{r}^{k}$ in Step~1 in increasing order of index $k$, $\forall~k\in \mathcal{S}$. 
Starting from $k=1$ to $S-1$, we iteratively replace $ct^{k}_{l_{max}}$ with $ct^{k}_{l_{max}} \bigwedge \Diamond(\phi^{k+1}_{r})$, where $ct^{k}_{l_{max}}$ is the last collaborative task in sorted $\widetilde{\mathbb{T}}^{k}_{r}$. Finally, $\phi_{r} = \phi_{r}^{1}$, and $\widetilde{\varphi}_{r} = \varphi_{r}\bigwedge \phi_{r}$. 

Note that the Step~$2$ actually imposes ordering constraints among independent subsequences, which seems to make the decomposition of the task sequence $\sigma$ in Sec.~\ref{3A} meaningless. However, we will show it is not the case.

\begin{remark}
\label{remark-decomposition}
	The decomposition of essential sequence in $\mathcal{F}$ reduces the amount of communication in real deployment.
For example, assume that there exists an essential sequence $\sigma = \{ct^{k_1}_1\}\{ct^{k_2}_1\}$ in which  $\sigma^{k_1}=\{ct^{k_1}_1\}$ and $\sigma^{k_2}=\{ct^{k_2}_1\}$ are two independent subsequences. 
If $\mathcal{R}(ct^{k_1}_1)\cap \mathcal{R}(ct^{k_2}_1)=\emptyset$, then the two tasks can be executed asynchronously. 
In contrast, without the decomposition of the essential sequence, all robots in $\mathcal{R}(ct^{k_2}_1)$ have to wait the completion of $ct^{k_1}_1$, which increases the total time cost.
\end{remark}

Given the $wTS_{r}$ and $\mathcal{F}_{r}$ of $\widetilde{\varphi}_{r}$, each robot $r$ can find its initial locally optimal run $\rho_{r}$ on a product automaton $\mathcal{P}_{r}$ as defined in Definition~\ref{pa}.

\begin{definition}[Product Automaton]
\label{pa}
	The product automaton of $wTS_{r}$ and $\mathcal{F}_{r}$ is a tuple $\mathcal{P}_{r}=wTS_{r}\otimes \mathcal{F}_{r} = \langle \mathcal{Q}_{P},\mathcal{Q}_{P}^{0},\rightarrow_{P}, \\ \omega_{P},\mathcal{Q}_{P}^{F} \rangle$, 
	where 
	$\mathcal{Q}_{P} = \mathcal{Q}_{r}\times \mathcal{Q}_{F}$;
	$\mathcal{Q}_{P}^{0} = q^{0}_{i}\times \mathcal{Q}_{F}^{0}$;  
	$\rightarrow_{P} \subseteq \mathcal{Q}_{P} \times \mathcal{Q}_{P}$, which satisfies that $\forall~(q_{P},q_{P}')\in \rightarrow_{P}$, it holds that
	$(\Pi_{r}q_{P}, \Pi_{r}q_{P}')\in \rightarrow_{i}$ and $L_{r}(\Pi_{r} q_{P}) \vDash \delta(\Pi_{F}q_{P}, \Pi_{F}q_{P}')$.
	Here $\Pi_{r}$ and $\Pi_{F}$ represent the projections into the state spaces of $wTS_{r}$ and $\mathcal{F}_{r}$, respectively;
	$\omega_{P}(q_{P},q_{P}')=\omega_{i}(\Pi_{r}q_{P},\Pi_{r}q_{P}')$;
	$\mathcal{Q}_{P}^{F}\subseteq \mathcal{Q}_{r}\times \mathcal{Q}_{F}^{F}$. 
\end{definition}

To identify the states in $\mathcal{P}_{r}$ related to the execution of collaborative tasks $ct\in \widetilde{\mathbb{T}}_{r}$, we define the collaborative state in Def.~\ref{collaborativestate}.

\begin{definition}[Collaborative State]
\label{collaborativestate}
For ${\forall}\ q_{P}\in \mathcal{Q}_{P}$, $q_{P}\in \mathcal{C}(ct)$ iff the following two conditions hold. Here $\mathcal{C}(ct)$ is a collection of all collaborative states in $\mathcal{P}_{r}$ of collaborative task $ct$. 
\begin{enumerate}
    \item $ct\in L_{r}(\Pi_{r}q_{P})$;
    \item ${\exists} q_{P}'\in \mathcal{Q}_{P}, q_{P} \in \delta(q_{P}', ct)$, and $\Pi_{\mathcal{F}}q_{P}' \ne \Pi_{\mathcal{F}}q_{P}$.
\end{enumerate}
\end{definition}

The second condition ensures that the robot plans to perform task $ct$ at state $q_{P}$ rather than just going through that state.

We utilize the Dijkstra algorithm to search for the shortest accepting run $\rho_{r}$ in $\mathcal{P}_{r}$ for each robot $r\in \mathcal{N}$.
The obtained optimal accepting run $\rho_{r}$ satisfies both task specifications and motion and task performing capabilities of robot $r$. 
The initial task execution strategy $\tau_r$ of robot $r$ can be obtained by projecting the run $\rho_{r}$ into the state space of $wTS_{r}$, i.e., $\tau_r = \Pi_{r} \rho_{r}$.

\textbf{Performing Execution Strategies:} Given the task execution strategy $\tau_r$ for each robot $r\in \N$, robot $r$ moves in the environment to perform tasks following the order in $\tau_r$. 
(1) If robot $r$ reaches a state $q_{P}\in \Q_P$ that holds $L_{r}(q_{P})\subseteq \overline{\T}_r$, it performs all tasks in $L_{r}(q_{P})$ individually before going to the next region;
(2) if robot $r$ reaches a state $q_{P}\in \Q_P$ and $\exists ct\in \tildeT_{r}, q_{P}\in \C(ct)$, then robot $r$ has to wait for the arrival of all other robots in $\R(ct)\backslash\{r\}$ and then perform the task together.
Once robot $r$ moves to the last state in $\tau_r$ and completes the potential tasks there, it completes its tasks.

When the robots execute the initially obtained task execution strategies, there may exist extra wait time in each collaboration, caused by different arrival times of robots to one collaborative task. 
This is because the robots synthesize their execution strategies independently without considering others.
In the next section, the initial task execution strategies will be optimized.

\begin{proposition}
\label{prop: completeness1}
(\textbf{Completeness})
If there exists a solution for Problem 1, then the proposed method can find task execution strategies $\{\tau_r\}_{r\in \mathcal{N}}$, when executed asynchronously and following the synchronization constraints in collaborations, can satisfy the individual task specification $\varphi_{r}$ for each robot $r\in \mathcal{N}$ and the global collaborative task specification $\phi$.
\end{proposition}

\begin{Proof}
If there exists a solution $\{\tau_r\}_{r\in \mathcal{N}}$ for Problem 1, it satisfies that:
(1) the joint behaviors of the robots w.r.t the collaborative tasks must follow an essential sequence $\sigma$ in the prune NFA $\mathcal{F}$ as in Sec.~\ref{3A} which describes an accepting run;
(2) the individual execution of each $\tau_r$ must induce an essential sequence of an accepting run $\rho_{r}$ in $\mathcal{F}_{r}$ of $\widetilde{\varphi}_r$ in Sec.~\ref{3c}, since $\mathcal{F}_{r}$ contains all possible task sequences to satisfy $\varphi_{r}$, and the constraints about the assigned collaborative tasks.
It is obvious that any existing solutions will also present in the searching space of the proposed method.

Then we prove that the proposed way for robots to perform their execution strategies will describe accepting runs on the NFAs of individual ${\rm LTL}_f$ task specifications and also on the NFA of collaborative ${\rm LTL}_f$ task specification. 
First, in the proposed framework, robots execute their execution strategies asynchronously and only synchronize with each other when performing collaborative tasks. 
In contrast, traditional method assumes synchronization of all robots in each step.
According to Def.~\ref{def-1}, robots only need to synchronize with each other when making the transition condition become true. 
During the robots moving between regions without performing actions, there is no enabled atom propositions.
This is also valid because we exclude the next operator from finite LTL, so there always exist a feasible self loop at each state in $\F_r$ whose transition condition always being true.

Second, in the induced task sequence while robot performing their strategies, the individual tasks and collaborative tasks may be performed alternately. This is also valid because we have assumed that the execution of individual task specification are independent with each other and also with the collaborative task specification, so when robots execute individual tasks, the state of collaborative specification remains on the self-loop, with no proposition of collaborative tasks enabled. The satisfaction of individual task specification is also similar to the collaborative specification.

To conclude, if there exists a solution to Problem $1$, it can be found by the proposed method. And performing the execution strategies by robots will satisfy both individual and collaborative task specifications.
\hfill~$\blacksquare$
\end{Proof}

 \begin{algorithm}[!t]
\setstretch{1}
\SetKwInOut{Input}{Input}\SetKwInOut{Output}{Output}
\SetKwInOut{Return}{Return}
\label{alg-framework}
\caption{Framework}
\small{
\Input{$\{\varphi_{r}\}_{r\in \mathcal{N}}$, $\phi$,  $\mathcal{W}$, $\tildeT$, $\{\overline{\mathbb{T}}_r\}_{r\in \N}$}
\Output{$\{\tau_{r}\}_{r\in \mathcal{N}}$}
Construct $wTS_{r}$ for $r\in \mathcal{N}$.\\
Convert $\phi$ to its corresponding NFA $\mathcal{F}$.\\
Prune $\mathcal{F}$ and identify the decomposition states in $\mathcal{F}$. Select and decompose the collaborative task execution sequence $\sigma = \sigma^{1} ;\sigma^{2};\dots;\sigma^{S}$ of an accepting run $\rho_{F}$ in $\mathcal{F}$.\\
Construct the SMT-based task allocation model $f$.\\
\While{$\textbf{{\rm SAT}}(f)$}{
    $\mathcal{X}\leftarrow \textbf{\rm SMT-solver}(f)$\\
    Pass $\mathcal{X}$ to each robot.\\
    \For{$r\in \mathcal{N}$}{
        Construct ${\rm LTL}_f$ formula $\phi_{r} \leftarrow$ that captures the assigned $ct\in \widetilde{\mathbb{T}}_r$ according to $\mathcal{X}$ and the temporal constraints\\
        $\widetilde{\varphi}_{r} \leftarrow \varphi_{r}\bigwedge \phi_{r}$.\\
        $\mathcal{P}_{r}\leftarrow wTS_{r}\bigotimes \mathcal{F}_{r}$~~~~~~~$//$  $\mathcal{F}_{r}$ is the NFA of $\widetilde{\mathcal{\varphi}}_r$\\
        $\rho_{r}\leftarrow$ local optimal path of $\mathcal{P}_{r}$\\
    }
    Construct pruned local product automaton $\LP_r$, $\forall r\in \N$.\\
    $\textbf{{\rm StrategyAdjustingMechanism}}\left(\{\LP_r\}_{r\in \mathcal{N}}, \{\rho_{r}\}_{r\in \mathcal{N}}\right )$\\
    $f\leftarrow f \bigwedge\neg X$\\
}
\KwRet{$\tau_{r}\leftarrow \Pi_{r}\rho_{r}$ for each robot $r\in\mathcal{N}$.}
}
\end{algorithm}

\section{Optimization of Task Execution Strategies}
\label{sec: optimization}

In this section, we propose two methods to optimize the initial execution strategies obtained in Sec.~\ref{3c}, given the assignments of collaborative tasks. 
We first prune each robot's local product automaton, so that only the states related to the execution of collaborative tasks are preserved.
Based on the pruned local product automaton, we then build a baseline MILP model to calculate the optimal execution strategies w.r.t the total time cost for robots.
Finally, we extend the execution strategy adjusting mechanism proposed in~\cite{individual-and-collaborative_arxiv_2021} by operating it on the pruned product automatons, and develop a distributed version to better distribute the computation burden. 
Experimental results about these two methods will be shown in Sec.~\ref{sec: experiment}.

\subsection{Pruning Local Product Automaton}
To optimize time efficiency of the obtained execution strategies, we only care about the states related to the execution of collaborative tasks. The reason is that the total time cost of robots depends on: (1) the time each robot spends to perform its execution strategies without considering wait time in collaborations; (2) the extra wait time each robot spends in each collaboration.

The pruned local product automaton of $\mathcal{P}_{r}$, denoted by $\LP_r$ only maintains three types of states: 1) initial states $q_P^0 \in \Q_P^0$; 2) accept states $q_P^F \in \Q_P^F$; 3) collaborative states $q\in \C(ct)$ for all $ct\in \widetilde{\mathbb{T}}_{r}$.
We define the pruned local product automaton as follows:
\begin{definition}  
\label{def: prune_local_product}
A pruned local product automation $\PP_r$ of robot $r$ is a tuple $\LP_r=\langle \Q_{\Lp}, \Q_{\Lp}^0, \rightarrow_{\Lp}, \omega_{\Lp}, \Q_{\Lp}^F \rangle$, 
where: 
\begin{enumerate}
    \item $\Q_{\Lp}\in \left(\bigcup_{ct\in \tildeT_{r}} \{\C(ct)\}\right)\bigcup \Q_{\Lp}^0 \bigcup \Q_{\Lp}^F$;
    \item $\Q_{\Lp}^0 = \Q_P^0$;
    \item $\rightarrow_{\Lp} \subseteq \Q_{\Lp}\times \Q_{\Lp}$, and $\forall\ (q_{\Lp}, q_{\Lp}')\in \rightarrow_{\Lp}$, there exists a path $\rho$ from $q_{\Lp}$ to $q_{\Lp}'$ in $\PP_i$. There are three types of edges in $\rightarrow_{\Lp}$. 
    $\forall (q_{\Lp}, q_{\Lp}')\in \rightarrow_{\Lp}$:
    \begin{enumerate}
        \item if $q_{\Lp}\in \C(ct), q_{\Lp}'\in \C(ct')$, $ct, ct'\in\tildeT_{r}$, then $ct$ must be the previous elements of $ct'$ in $\tildeT_{r}$;
        \item else, if $q_{\Lp}\in \Q_{\Lp}^0$ and $q_{\Lp}'\in \C(ct)$, then $ct$ must be the first element in $\tildeT_{r}$;
        \item else, if $q_{\Lp}\in \C(ct)$ and $q_{\Lp}'\in \Q_{\Lp}^F$, then $ct$ must be the last element in $\tildeT_{r}$.
    \end{enumerate}
    \item $\omega_{\Lp}(q_{\Lp}, q_{\Lp}') = \sum_{k = 1}^{|\rho|-1}\omega_P(\rho(k), \rho(k-1)$, here $\rho$ is the shortest directed path connecting $q_{\Lp}$ and $q_{\Lp}'$ in $\PP_r$;
    \item $\Q_{\Lp}^F = \Q_P^F$.
\end{enumerate}

The pruned product automaton $\LP_r$ has a hierarchy structure, in which each level contains all collaborative states corresponds to one specific collaborative tasks in $\tildeT_{r}$ (except the first and the last levels), and the order of each level follows the order of corresponding tasks in $\tildeT_{r}$.

\begin{proposition}
\label{prop_prunecomplexity}
The pruned local product automaton $\LP_r$ can be constructed in $O(\Delta^{2}\cdot |\tildeT_r| \cdot (E\cdot lg E))$, where $\Delta$ is the magnitude of collaborative states in $\PP_r$ for each collaborative task $ct\in \tildeT_r$, and $E\cdot lg E$ is the time complexity of using Dijkstra algorithm on $\LP_r$ to search for the shortest path from $q_{\Lp}$ to $q_{\Lp}'$ for an edge $(q_{\Lp}, q_{\Lp}')$ in Def.~\ref{def: prune_local_product}, and $E$ is the number of edges in $\PP_r$.
\end{proposition}

\end{definition}

\subsection{MILP-based Baseline Method}

In this section, we aim to find the optimal execution strategies for robots to satisfy all task requirements while minimizing the total time cost.
We formalize the optimization problem as an MILP model. 
The solution to the MILP model is a set of accepting paths on robots' pruned local product automatons, from which the execution strategies for robots can then be extracted.
The MILP model can be solved by off-the-shell solving tools like Gurobi~\cite{gurobi}. 

The MILP model is based on the pruned local product automaton. 
First, we define a set of Boolean variable $\Y_r = \{y_r^{(i, j)} | (q_i, q_j)\in \rightarrow_{\Lp}\}$ for each robot $r\in \N$.
The variable $y_r^{(i, j)}$ is true indicating that the corresponding edge $(q_i, q_j)$ is selected in the accepting run of robot $r$.
The final execution strategy of robot $r$ can be generated according to the value of $\Y_r$.
The values of variables in $\Y_r$ are guaranteed to generate a feasible accepting run on $\LP_r$ by applying following constraints:
\begin{align}
    & y^{(i, j)}_r = 1\ or\ 0, \forall~(q_i, q_j)\in \rightarrow_{\Lp}, r\in \N;\\
    & \sum_{q_k\in Pre(i)}y^{(k, i)}_r = \sum_{q_j\in Sub(i)}y^{(i, j)}_r \le 1, \forall~q_i\in \Q_{\Lp}\backslash (Q^0_{\Lp} \cap Q^F_{\Lp});\\
    & \sum_{q_i\in \Q_{\Lp}^0}\sum_{q_j\in Sub(i)}y^{(i, j)}_r = 1;\\
    & \sum_{q_i\in \Q_{\Lp}^F}\sum_{q_k\in Pre(i)}y^{(k, i)}_r = 1.
\end{align}

Here $Pre(i)$ and $Sub(i)$ denote the set of predecessors and successors of state $q_i$ in $\LP_r$ respectively.

By applying the above constraints $(1)-(4)$, the variables $y_r^{(i, j)}\in \Y_r$ that take true can generate an accepting path on the prune local product automaton $\LP_r$.

Then we formalize the wait time of robots in collaborations.
For each $ct^k_l\in \tildeT_r$, we define two types of auxiliary variables $t^{(k, l)}_{r}$ and $delay^{(k, l)}_{r}$ respectively. 
The variable $delay^{(k, l)}_r$ equals to the total wait time for robot $r$ after the execution of $ct^k_l$.
Initially, all these variables are assigned $0$.
We have 
\begin{equation}
    t^{(k, l)}_{r} = \sum_{i\in Pre(j)}y_{r}^{(i, j)}\cdot (t^{(k', l')}_{r} + \omega_{\Lp}(q_i, q_j)),
\end{equation}
 where $ct^{k'}_{l'}$ is the previous task of $ct^k_l$ in $\tildeT_r$, and $q_i \in \C(ct^{k'}_{l'})$, $q_j\in \C(ct^{k}_{l})$. 
 According to the Def.~\ref{def: prune_local_product}, there exists an edge $(q_i, q_j)\in \rightarrow_{\Lp}$, and $\omega_{\Lp}(q_i, q_j)$ is the transition time between the two states.
The variable $t^{(k, l)}_{r}$ actually represents robot $r$' ideal arrival time to the region of task $ct^k_l$.
Note that for the first element in $\tildeT_r$, it has no previous element, so it only takes the second term $\omega_{\Lp}(q_i, q_j)$ into account.

We define a set of variables $\Z = \{z^k_l | \exists ct^k_l\in \tildeT, z^k_l =\max_{r\in \N}(t^{(k,l)}_{r} + delay^{(k', l')}_r)\}$ to represent the latest arrival time for each collaborative task.
Here $delay^{(k', l')}_r$ is the total wait time for robot $r$ after the execution of $ct^{k'}_{l'}$.
We have the constraints that:
\begin{equation}
    delay^{(k, l)}_r = z^k_l - t^{(k, l)}_{r}, \forall ct^k_l\in \T_r.
\end{equation}

Finally, the objective function is defined as:
\begin{equation}
    J = \sum_{r\in \N}t_{r}^{(k_e, l_e)} + delay_r^{(k_e, l_e)},
\end{equation}
where $k_e$, $l_e$ corresponds to $ct^{k_e}_{l_e}$, the last element in $\tildeT_r$.

After solving the above MILP model, we can get each robot's execution strategy according to the value of variables in $\X_r$.

\textbf{Time Complexity:} 
the number of variables in the MILP model is $O(N\cdot \Delta^{2}\cdot |\tildeT_r| + |\tildeT|)$.
For each robot $r\in \N$, the hierarchy structured pruned local product automaton $\LP_r$ has $O(|\tildeT_r|)$ levels.
Each level contains collaborative states for one $ct$ in $\tildeT_r$ of magnitude $\Delta$ in $\PP_r$.
The number of edges in $\LP_r$ is $O(\Delta^{2}|\tildeT_r|)$, and so is the number of variables in $\Y_r$.
Additionally, the number of variables $t_{r}^{(k, l)}$ and $delay_{r}^{(k, l)}$ equals to the number of $\tildeT_r$, and the number of variable $\Z$ equals to $\tildeT$.

\subsection{Iterative Execution Strategy Adjusting Mechanism}

\begin{algorithm}[!t]
\setstretch{0.8}
\SetKwInOut{Input}{Input}\SetKwInOut{Output}{Output}
\SetKwFunction{optimizeTime}{optimizeTime}
\label{alg_mechanism}
\caption{StrategyAdjustingMechanism}

\texttt{Initialize:}\\
$count \leftarrow 0$, $ct \leftarrow$ the first task in $\tildeT^{sort}$.\\
$tl_r \leftarrow$ ComputeTimeline($r$), for $r\in \N$.\\
Each robot $r\in \N$ propagates $tl_r$ to other robots.\\
$msg \leftarrow \langle False, \_, \_, \_, ct, count\rangle$.\\
Propagate $msg$ to all robots.\\

~\\

\texttt{when robot r received(msg):}\\
    $\langle success, r', received, tl_{r'}, ct, count\rangle \leftarrow msg$.\\
    \If{success}{
        Modify robot $r$'s $tl_{r'}$ to the received $tl_{r'}$.\\
    }
    \eIf{$r\notin received \wedge r\in \R(ct) \wedge r={\rm FindLatest}$$(ct)$}{
        Add robot $r$ into $received$.\\
        $\canopt \leftarrow$ AdjustStrategy($ct$, $isLatest$ = true).\\
        $ct^{next}\leftarrow$ the next task of ct in $\tildeT^{sort}$.\\
        \eIf{\canopt}{
            $tl_r \leftarrow$ ComputeTimeline($r$).\\
            $msg \leftarrow \langle canOpt, r, \{\}, tl_r, ct^{next}, count+1\rangle$.\\
            propagate $msg$ to all other robots.\\
        }{
        $r'\leftarrow$ FindEarliest($ct$).\\
        $token \leftarrow \langle {r}, r', msg\rangle$.\\
        Propagate $token$ to $r'$, the earliest robot.\\
        }
    }{
        send $msg$ to other robots not in $received$.\\
    }
~\\
\texttt{when robot r received(token):}\\
    $\langle r, r', msg\rangle \leftarrow token$.\\
    \eIf{$r = r'$}{
        \canopt $\leftarrow$ AdjustStrategy($ct$, $isLatest$=false).\\
        $ct^{next}\leftarrow$ the next task of ct in $\tildeT^{sort}$.\\
        \eIf{\canopt}{
            $count \leftarrow count+1$.\\
            $tl_r\leftarrow$ ComputeTimeline($r$).\\
        }{
            \If{$ct$ is the last task}{
                \eIf{$count = 0$}{
                    Send ternimation to other robots.\\
                    \textbf{Ternimate}.\\
                }
                {   
                    $count \leftarrow 0$, $ct^{next}\leftarrow$ the first task in $\tildeT^{sort}$.\\
                }
            }
        }
        $msg \leftarrow \langle canOpt, r, \{\}, tl_r, ct^{next}, count\rangle$.\\
        propagate $msg$ to all other robots.\\
    }{
        received.add(r).\\
        Propagate $token$ to $r'$.\\
    }
~\\
\SetKwFunction{FMain}{FindLatest}
\SetKwProg{Fn}{Function}{:}{}
\SetKwInOut{Input}{Input}\SetKwInOut{Output}{Output}
\SetKwInOut{Return}{Return}
\Fn{\FMain{$ct$}}{
    \small{
        $ct^{pre}\leftarrow$ the previous task of $ct$ in $\tildeT_{r}$.\\
        $r \leftarrow \arg \max\limits_{r\in \mathcal{R}(ct)}\left\{t(ct^{pre})-t_{r}(ct^{pre}) + t_{r}(ct)\right\}$.\\
        \KwRet $r$\\
    }
}    
~\\
\SetKwFunction{FMain}{FindEarliest}
\SetKwProg{Fn}{Function}{:}{}
\SetKwInOut{Input}{Input}\SetKwInOut{Output}{Output}
\SetKwInOut{Return}{Return}
\Fn{\FMain{$ct$}}{
    \small{
        $ct^{pre}\leftarrow$ the previous task of $ct$ in $\tildeT_{r}$.\\
        $r \leftarrow \arg \min\limits_{r\in \mathcal{R}(ct)}\left\{t(ct^{pre})-t_{r}(ct^{pre}) + t_{r}(ct)\right\}$.\\
        \KwRet $r$\\
    }
}
\end{algorithm}

\begin{algorithm}[!t]
\setstretch{1}
\SetKwInOut{Input}{Input}\SetKwInOut{Output}{Output}
\label{alg_timeline}
\caption{ComputeTimeline($r$)}
\small{
$tl_r = [~]$.\\
\For{$ct\in \tildeT_r$}{
$t_{r}(ct) = \sum_{j = 1}^{j^{ct}-1}\omega_{r}(\rho_{r}(j), \rho_{r}(j+1))$.\\
$tl_{r}.{\rm append}(t_{r}(ct))$.\\
}
$T_r = \sum_{j = 1}^{|\rho_r|-1}\omega_{r}(\rho_{r}(j), \rho_{r}(j+1))$.\\
$tl_{r}.{\rm append}(T_r)$.\\
\KwRet{$tl_{r}$}
}
\end{algorithm}

\begin{algorithm}[!t]
\setstretch{1}
\SetKwInOut{Input}{Input}\SetKwInOut{Output}{Output}
\label{alg_timecost}
\caption{ComputeTimeCost}
\small{
 ${\rm delay}_{r} \leftarrow 0$, for $r \in \mathcal{N}$.\\
 \For{$ct\in \widetilde{\mathbb{T}}^{sort}$}{
    $\hat{t}_{r}(ct) = t_{r}(ct) + delay_{r}$, for $r\in \R(ct)$.\\
    $t(ct) = \max_{r\in \mathcal{R}(ct)}\hat{t}_{r}(ct)$.\\
    ${\rm delay}_{r} = t(ct) - t_{r}(ct)$, for $r\in \mathcal{R}(ct)$.\\
 }
 $T_{r}^{\rm colla}\leftarrow T_{r} + {\rm delay}_{r}$, for $r\in \mathcal{N}$.\\
 $T^{\rm colla} = \sum_{r\in \mathcal{N}}T_{r}^{\rm colla}$.\\
\KwRet{$T^{\rm colla}$}
}
\end{algorithm}

\begin{figure}[t]
    \centering
    \includegraphics[width=\linewidth]{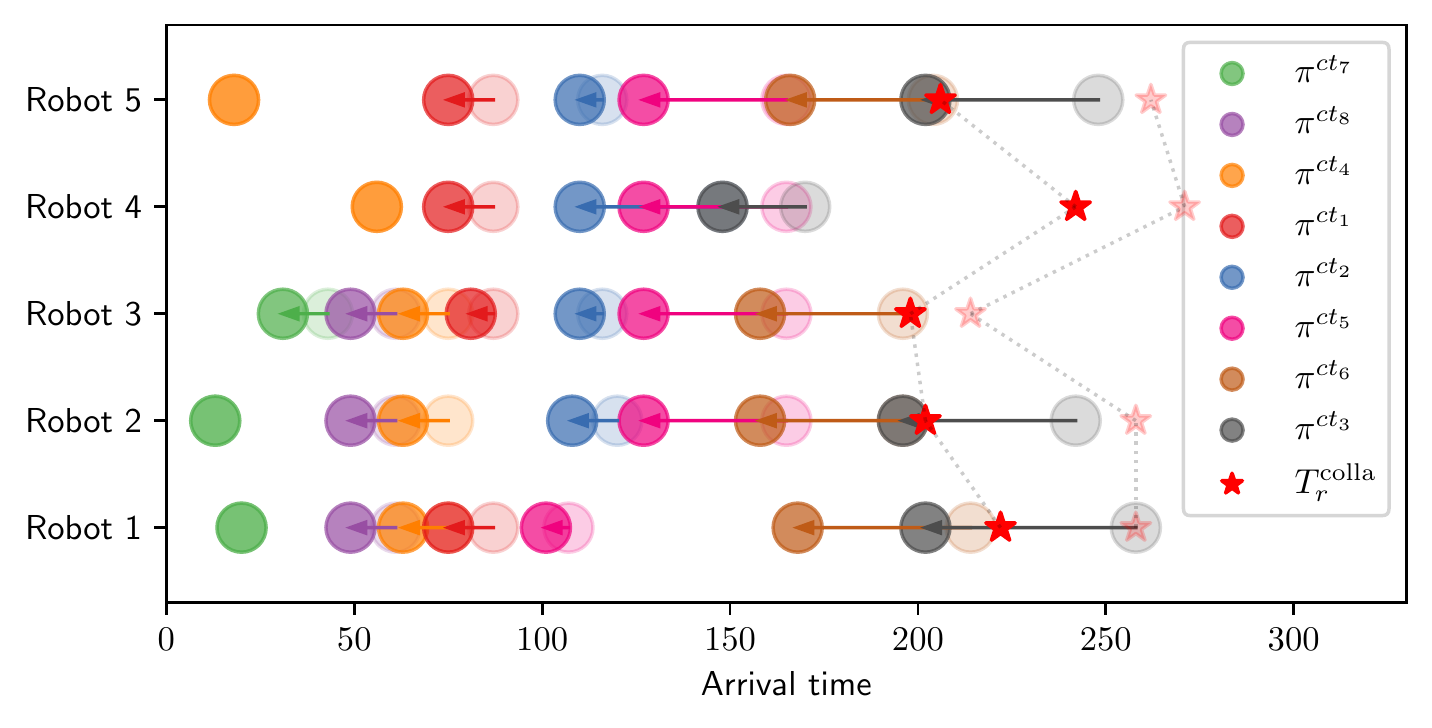}
    \caption{Robots' arrival time to the assigned collaborative tasks before (transparent discs) and after (solid discs) applying the execution strategy adjusting mechanism in an experiment with $5$ robots.
    The red stars represent each robot's time cost before (transparent) and after (solid) the adjusting while accounting the potential wait time in collaborations.
    The collaborative ${\rm LTL}_f$ formula is $\phi = (\Diamond (\pi^{ct_1} \wedge (\Diamond \pi^{ct_2}))) \wedge (\Diamond \pi^{ct_3}) \wedge (\Diamond \pi^{ct_4}) \wedge (\neg \pi^{ct_1}~{\rm U}~(\Diamond \pi^{ct_6})) \wedge (\Diamond \pi^{ct_7}) \wedge (\Diamond \pi^{ct_8}) \wedge (\neg \pi^{ct_5}~{\rm U}~\pi^{ct_8})$.
    The figure illustrates the effect of the proposed adjusting mechanism in reducing total time cost.}
    \label{fig:adj_vs_ip}
\end{figure}

In this subsection, we propose an distributed execution strategy adjusting mechanism to iteratively optimize the total time cost, in which the computational burden is distributed among robots.
A preliminary mechanism has been introduced in \cite{individual-and-collaborative_arxiv_2021}, and in this paper, we further improve the method by operating on the pruned local product automaton, which can greatly improve its solving efficiency, as shown in Sec.~\ref{sec: experiment}.

Intuitively, the adjusting mechanism investigates whether the total time cost can be reduced by adjusting robots' arrival time to regions of collaboration. The adjusting is inspired by reducing inefficient wait time in each collaborations.
Fig.~\ref{alg_timeline} shows robots' arrival time to the assigned collaborative tasks before and after applying the proposed adjusting mechanism.
The proposed execution strategy adjusting mechanism operates in a distributed manner, as described in Alg.~\ref{alg_mechanism}. 
We assume that the communication network between the robots are connected, so that messages can spread over all robots.
We will explain the details of the adjusting mechanism in the following context.

\subsubsection{\textbf{Initialization}}
In this procedure, each robot computes a timeline and propagate it to all other robots.
Let $tl_r$ denotes the timeline of robot $r$, which can be constructed according to Alg.~\ref{alg_timeline}.
The timeline $tl_r$ records robot $r$'s ideal arrival time to the region of the assigned collaborative tasks according to its current execution strategy $\rho_r$, without considering the potential wait time. 
It also includes the time robot $r$ spends to finish all its tasks, also without considering wait time (Line $5-6$, Alg.~\ref{alg_timeline}).
After robot $r$ receiving $tl_{r'}$ from all other robots $r'\in \N\backslash\{r\}$, it can locally compute the actual execution time of the assigned collaborative tasks and the total time cost, considering the potential synchronization constraints with other robots, as shown in Alg.~\ref{alg_timecost}.
In Alg.~\ref{alg_timecost}, each robot can obtain the exact arrival time of each robot to one collaborative task in $\tildeT^{sort}$, i.e., $\hat{t}_{r}(ct)$ denotes the time when robot $r$ arrives at the region of $ct$, including previous time $delay_r$ due to the synchronization constraints in collaborations.
Note that $delay_r$ is iteratively modified as the execution of collaborative tasks. 
Moreover, the Alg.~\ref{alg_timecost} can also calculate the total time cost for all robots, denoted as $T^{colla}$ (Line $6-7$), which will be used in the adjusting procedure to check whether a modified execution strategy can result in a better performance.

Additionally, the central server will generate a message $msg$ and propagate it to all robots, which will start the adjusting procedure.
More specifically, this $msg$ will make the last robot arriving to the first collaborative task automatically get the token. 
The formal definition of the message $msg$ will be given in next subsection.

\subsubsection{\textbf{Adjusting Procedure}}

In this procedure, the robots traverse all collaborative tasks to investigate whether the wait time caused by synchronization constraints and the total time cost can be reduced by adjusting their execution strategy locally.
The procedure depends on the propagation of two types of messages defined as follows.
\begin{itemize}
    \item $msg$, the broadcast message. A $msg$ is a tuple $\langle canOpt, r, received, tl_r, ct, count\rangle$, 
where $canOpt$ is a Boolean variable indicating whether the previous robot $r$ optimizes the total time cost; 
$tl_r$ is the modified timeline for robot $r$;
$ct$ is the current collaborative task to be investigated and $count$ counts the times of successful optimization of the total time cost within one cycle.
    \item $token$, the directional message. A $token$ is a tuple $\langle r, r', msg \rangle$, where $r$ and $r'$ are source and target robot respectively.
\end{itemize}

When a robot $r$ receives an $msg$ from a robot $r'$ (as the case after the initialization procedure), it checks whether it is the latest robot participating into the collaboration of task $ct$ by calling function \texttt{FindLatest()} (Line $12$, Alg.~\ref{alg_mechanism}).
If it is not the case, robot $r$ adds itself into the set $received$ of the $msg$, and sends the $msg$ to all other robots not in $received$ through communication network.
If it is the case, then robot $r$ starts to adjust its execution strategy, as described in Alg.~\ref{alg_adjust}.
The detailed will be explained later.
The Alg.~\ref{alg_adjust} returns a Boolean value $canOpt$, indicating whether robot $r$ can find another execution strategy so that the total time cost can be optimized.
If $canOpt$ is true, then a new $msg$ is constructed, which includes the modified $tl_r$ according to robot $r$'s new execution strategy (Line $16-19$, Alg.~\ref{alg_mechanism}).

Otherwise, if $canOpt$ is false, robot $r$ finds the earliest robot arriving to the current collaborative task $ct$ by calling function \texttt{FindEarliest()}.
Then robot $r$ construct a message $token = \langle r, r', msg\rangle$ and send it to robot $r'$ through the communication network.
If in a special case $r = r'$, then a message $msg$ is directly constructed and is sent to all other robots.
Note that although current robot $r$ can find out which robot should get the token for task $ct^{next}$, the $msg$ still need to be propagated to all other robots to update their local $tl_r$.

When a robot $r$ receives a $token$, i.e, it is the earliest robot arriving to current collaborative task, it has the authority to adjust its execution strategy, trying to reduce its wait time in current collaboration and the total time cost (Line $29-30$, Alg.~\ref{alg_mechanism}.
The adjusting process is similar to what the latest robot does. 
The difference is, if the total time cost cannot be optimized, the token will be directly passed for the next collaborative task in $\tildeT^{sort}$.

\begin{algorithm}[!t]
\setstretch{0.8}
\SetKwInOut{Input}{Input}\SetKwInOut{Output}{Output}
\SetKwFunction{optimizeTime}{optimizeTime}
\label{alg_adjust}
\caption{AdjustStrategy($ct$, $isLatest$)}
\small{
$ct^{pre}\leftarrow$ the previous task of $ct$ in $\widetilde{\mathbb{T}}_{r}$.\\
\For{$q\in \mathcal{C}(ct)\backslash\{\rho_{r}(j^{ct})\}$}{
    $\rho'_{r} \leftarrow$ $\rho_{r}[..j^{pre}-1]$
    + Dijkstra$\left(\mathcal{P}_{r}, \rho_{r}(j^{pre}), q\right)$ \\ ~~~~~~~~+Dijkstra$(\mathcal{P}_{r}, q, \mathcal{Q}_{P}^{F})$\\
    \If{\parbox[t]{1\linewidth}{$\left(isLatest \bigwedge t(ct^{pre})-t_{r}(ct^{pre})+t'_{r}(ct)<t_{r}(ct)\right)$\\
    ${\rm or}$ $\left((\neg~isLatest)\bigwedge\right.$\\ $\left.t_r(ct) < t(ct^{pre})-t_{r}(ct^{pre})+t'_{r}(ct)\le t(ct)\right)$}}{
        $T^{\rm colla}_{\rm cand} \leftarrow\ $ComputeTimeCost$(\rho'_{r})$\\
        \If{$T^{\rm colla}_{\rm cand}<T^{\rm colla}$}{
            $\rho_{r} \leftarrow \rho'_{r}$\\
            \KwRet true\\
        }
    }
}
\KwRet false\\
}
\end{algorithm}

Furthermore, the detailed adjusting strategy is explained in Alg.~\ref{alg_adjust}.
Assume that robot $r$ takes the token to adjust its execution strategy for one task $ct\in \tildeT_r$.
The set $\mathcal{C}(ct)$ of $\mathcal{P}_{r}$ actually provides all candidates states corresponds to the execution of task $ct$, namely, all possible arrival time instances to $ct$.
The robot $r$ randomly traverses all state $q\in \mathcal{C}(ct)\backslash\{\rho_{r}(j^{ct})\}$ until finding one candidate state $q$ which can optimize $T^{\rm colla}$.
For each candidate state $q$, robot $r$ searches a candidate run $\rho'_{r}$ in $\mathcal{P}_{r}$: first maintain the original $\rho_r$ unchanged until $\rho_r(j^{pre}-1)$; then concatenate a new shortest path starting from $\rho_{r}(j^{prev})$ to the accepting states of $\mathcal{P}_{r}$, while forcing the path to pass through $q$ (Line $3-4$).
Here $\rho_{r}(j^{prev})$ is the collaborative state of task $ct^{prev}$ in $\rho_{r}$ and $ct^{prev}$ is the previous collaborative task of $ct$ in $\widetilde{\mathbb{T}}_{r}$ (Line $2$).
The candidate run $\rho'_{r}$ is selected to be the new $\rho_{r}$ if the following two conditions hold (Lines~$5-8$): 
\begin{enumerate}
    \item[(a)] if robot $r$ is the latest robot, it holds that $t(ct^{prev})-t_{r}(ct^{prev})+t'_{r}(ct)<t_{r}(ct)$; 
    
    if robot $r$ is the earliest robot, it holds that $t_i(ct) < t(ct^{prev})-t_{r}(ct^{prev})+t'_{r}(ct)\le t(ct)$.
    \item[(b)] $T^{\rm colla}_{\rm cand}<T^{\rm colla}$, here $T^{\rm colla}_{\rm cand}$ is the total time cost after replacing $\rho_r$ by $\rho_{r}'$.
\end{enumerate}
The condition (a) ensures that in the candidate run $\rho'_{r}$, the time robot $r$, the latest robot arriving to $ct$, is advanced (or delayed, for the earliest robot), such that the wait time for other robots in the collaboration of task $ct$ is reduced. 
Here $t'_i(ct)$ is calculated from the candidate run $\rho'_{r}$.
The condition (b) further guarantees the above local greedily adjustment procedure will also contribute to the total time cost $T^{\rm colla}$.
If robot $r$ cannot find a qualified candidate run $\rho'_{r}$ that satisfies the above two conditions, then the algorithm will return false.

\subsubsection{\textbf{Termination}}
If in one iteration no optimization happens, i.e., $count = 0$, then the adjusting procedure will terminate. 
Otherwise, $count > 0$ means the modified execution strategy results in better total time cost $T^{\rm colla}$, and the adjusting procedure will go on to the next iteration to further optimize current strategies.

The above condition will be checked by the earliest robot arriving to the last task in $\tildeT^{sort}$ if it cannot optimize the total time cost after adjusting its execution strategy.
If the termination condition holds, then the robot will send the termination signal to all other robots to stop the adjusting procedure (Line $36-39$, Alg.~\ref{alg_mechanism}).

\begin{remark}
The proposed adjusting mechanism generates non-trivial execution strategies (trivial means the robot may pass through some unnecessary regions to deliberately postpone the arrival time in the collaboration), because we search the shortest path passing through the collaborative state which corresponds to the actual execution of tasks as in Def.~\ref{collaborativestate}.
\end{remark}

\subsection{Completeness, Optimality and Complexity}

\begin{proposition}
The proposed framework with execution strategy adjusting mechanism is complete.
\end{proposition}

\begin{Proof}
The proof follows Prop.~\ref{prop: completeness1}. Note that in the adjusting procedure, robots may select another accepting runs on their pruned local product automatons, which also has a corresponding run on the original local product automaton.
Such change does not violate Prop.~\ref{prop: completeness1}, so the proposed framework with execution strategy adjusting mechanism is also complete under the same assumptions as in Prop.~\ref{prop: completeness1}.
\hfill~$\blacksquare$
\end{Proof}

\begin{proposition}
\label{prop_monotonic}
The total time cost decreases monotonically when applying the execution strategy adjusting mechanism and the adjusting procedure terminates within finite iterations. 
\end{proposition}

\begin{Proof}
Given the collaborative task assignments, $T^{\rm colla}$ decreases monotonically from $T^{\rm colla}_{\rm init}$ until convergence, where $T^{\rm colla}_{\rm init}$ is the total time cost of initial execution strategies before adjusting, according to Lines $5-8$ of Alg.~\ref{alg_adjust}.
In addition, $T^{\rm colla}$ will not be smaller than $T^{\rm indiv}$, which is the ideal time cost of robots' initial execution strategies without accounting wait time.
Considering the resolution of discretization of the environment is limited, we conclude that Algorithm~\ref{alg_adjust} will terminate within finite iterations.
\hfill~$\blacksquare$
\end{Proof}

\begin{proposition}
The worst case time complexity of Alg.~\ref{alg_adjust} is $\mathcal{O}\left((T^{\rm colla}_{\rm init}-T^{\rm indiv})\cdot |\LP_{r}|\cdot(E\cdot \lg E + |\sigma|\cdot N)\right)$, where $N$ is the number of robots, $|\LP_{r}|$ is the maximum number of states of the local product automaton and $E$ is the maximum number of edges in $\LP_{r}$. 
$|\sigma|$ is the number of all distinct collaborative tasks in the selected accepting run $\sigma$ in Sec.~\ref{3A}.
\end{proposition}

\begin{Proof}
In Alg.~\ref{alg_adjust}, the robot with token will traverse all candidate collaborative states in $\LP_{r}$ in each adjustment in the worst case (Line $2$). 
The number of adjustments is limited by $\mathcal{O}(T^{\rm colla}_{\rm init}-T^{\rm indiv})$ as in Prop.~\ref{prop_monotonic}.
For each candidate collaborative state, the robot with token utilizes Dijkstra algorithm to search a run $\rho'_r$ with complexity $\mathcal{O}(E\cdot \lg E)$, and also calls Alg.~\ref{alg_timecost} to validate the candidate run with time complexity $|\sigma|\cdot N$.
To summarize, the time complexity of Alg.~\ref{alg_adjust} is $\mathcal{O}\left((T^{\rm colla}_{\rm init}-T^{\rm indiv})\cdot |\LP_{r}|\cdot(E\cdot \lg E + |\sigma|\cdot N)\right)$.
\hfill~$\blacksquare$
\end{Proof}

\textbf{Optimality: } The proposed framework does not guarantee to find the optimal results, because in the adjusting mechanism, the robots greedily adjust their task execution strategies, rather than exhaustively investigating the combinations of all possible candidates. 
The MILP formulation can obtain the optimal execution strategies given the assignment of collaborative tasks, which scales poorly as the number of robots increases (see experimental results in Sec.~\ref{sec: experiment}).
Moreover, the selection of the collaborative task sequence $\sigma$ in Sec.~\ref{3A} also influences the performance of the final results.

\section{Experiment}
\label{sec: experiment}

In this section, we evaluate the scalability, solution quality and efficiency of the proposed method. 
All the experiments are performed on a Ubuntu $16.04$ server with CPU Intel Xeon E$5$-$2660$ at $2.00$GHz and $128$ GB of RAM. We use Z3~\cite{10.1007/978-3-540-78800-3_24} SMT solver to solve the SMT formulation of collaborative task allocation problem.
The baseline MILP method is solved by Gurobi v$9$ \cite{gurobi}.
All the programs are written in Python~$3$, and we use the package $igraph$\footnote{https://igraph.org/python/} to support our operations on graphs, which is implemented in C language.

We assume a grid map environment where some local tasks and collaborative tasks are randomly distributed. Each robot $r$ has a local task specification $\varphi_{r}$, e.g., $\varphi_{r} = (\Diamond \pi^{ts_{1}}) \wedge (\Diamond \pi^{ts_{2}}) \wedge (\Diamond \pi^{ts_{3}}) \wedge (\Diamond \pi^{ts_{4}})\wedge (\neg \pi^{ts_{1}}~{\rm U}~\pi^{ts_{4}})$. 
Additionally, the team of robots needs to collaborate to satisfy collaborative task specification $\phi$, e.g., $\phi = (\Diamond \pi^{ct_{1}}) \wedge (\Diamond \pi^{ct_{2}}) \wedge (\Diamond \pi^{ct_{4}}) \wedge (\neg \pi^{ct_{3}}~{\rm U}~\pi^{ct_{2}}) \wedge (\Diamond (\pi^{ct_{4}} \wedge (\Diamond \pi^{ct_{3}})))$, which has $|\tildeT|=4$.
In our simulations, we assume that the completion of each $ct$ in $\widetilde{\mathbb{T}}$ requires collaboration of several robots having different capabilities from a set $\{c_{1}, c_{2}, c_{3}\}$.
The type and amount of robots needed for each collaborative task are randomly generated in the simulation.

\subsection{Scalability of the Framework} 

\begin{table}[]
\setlength{\abovecaptionskip}{0.1cm}
\caption{Baseline Method (B.S.) Vs The proposed Method (Proposed)}
\label{tab: compare_product}
\renewcommand{\arraystretch}{1.3}
\setlength{\tabcolsep}{1mm}{
\footnotesize{
\begin{center}
\begin{tabular}{|c|c|c|c|c|c|}
\hline
\multirow{2}{*}{N} & \multirow{2}{*}{$|\tildeT|$} & \multirow{2}{*}{$|\ddot{\mathcal{P}}|$} & \multirow{2}{*}{B.S.(sec/sec)} & \multicolumn{2}{c|}{Proposed(sec/sec)} \\ \cline{5-6} 
                   &                              &                                         &                                & first              & best              \\ \hline
\multirow{2}{*}{2} & 4                            & 34848                                   & 266/12.4                       & 252/11.5           & 252/47.5          \\ \cline{2-6} 
                   & 6                            & 161376                                  & 338/97.0                       & 352/28.2           & 352/57.6          \\ \hline
\multirow{2}{*}{3} & 4                            & 3241792                                 & 333/41962.6                    & 342/19.8           & 330/195.1         \\ \cline{2-6} 
                   & 6                            & $1.7\times 10^{7}$                      & -/-                            & 390/66.2           & 326/1800          \\ \hline
5                  & 6                            & $1.4\times 10^{11}$                     & -/-                            & 819/52.9           & 802/1800          \\ \hline
50                 & 6                            & $5.2\times 10^{97}$                     & -/-                            & 13702/700.4        & 13702/1800        \\ \hline
100                & 6                            & $2.7\times 10^{195}$                    & -/-                            & 26514/3548.7       & 26514/3548.7      \\ \hline
\end{tabular}
\end{center}
\vspace{0.3ex}
{\justifying 
The items in column B.S. and Proposed represent:  $T^{\rm colla}/t^{\rm cal}$, where $T^{\rm colla}$ is the total time cost and $t^{\rm cal}$ is the solving time. 
The ``-'' items indicate memory overflow. \par}
}}
\vspace{1ex}
\end{table}

We compare the proposed method with the traditional method based on the product automaton with state pruning techniques, which is similar to \cite{tumovaDecompositionMultiagentPlanning2015}. 
More specifically, each robot $r$ constructs a local product automaton $\dot{\mathcal{P}}_r=wTS_{r}\otimes \dot{\mathcal{F}}_i$, where the insignificant states in $wTS_r$ are pruned to maintain only the states corresponding to the task regions. $\dot{\mathcal{F}}_i$ is the NFA of $\varphi_{r}$.
The we build a global product automaton $\ddot{\mathcal{P}}=\dot{\mathcal{P}}\otimes \mathcal{F}$, where $\dot{\mathcal{P}}=\dot{\mathcal{P}}_1\otimes \cdots \otimes \dot{\mathcal{P}}_N$ and $\mathcal{F}$ corresponds to $\phi$.
Here the $\dot{\mathcal{F}}$, $\dot{\mathcal{P}}$ and $\ddot{\mathcal{P}}$ are used to distinguish from the notations used in our proposed method.
The task execution strategy can be obtained by searching the shortest accepting run in $\ddot{\mathcal{P}}$.
Note that the shortest accepting run may not be optimal w.r.t the total time cost, because the searching process doesn't take the potential wait time into account.
It remains a problem about how to efficiently find a path on above automaton $\ddot{\mathcal{P}}$ while also considering the potential wait time in collaborations, and here we just take the shortest path.
This only has little impact on our evaluation of the scalability of the proposed method.

We investigate the performance of the two methods in different task requirements and robot numbers, and keep the environment size as $30\times 30$ in all cases, as shown in Tab.~\ref{tab: compare_product}.
For the proposed framework, we record the time when obtaining the first feasible solutions and the best solutions within $30$ minutes.
The baseline method based on global product automaton quickly becomes intractable as the number of robots $N$ and collaborative tasks $|\tildeT|$ increases. 
Roughly speaking, the computational time of our proposed framework grows proportional to the number of robots. This is mainly because we accumulate the time spent by all robots to construct their local product automatons.
We have to point out that our method runs much quicker when the local planning procedure is distributed to individual robots, and scales well with large number of robots.
The results illustrate that the proposed method can quickly generate feasible execution strategies for the robots, and has high scalability compared with the baseline method.


\subsection{Optimization Efficiency of the Adjusting Mechanism}

\begin{table*}[]
\centering
\caption{Comparison of Computational Time for Pruning product automatons, Adjusting Mechanism and MILP Method}
\label{table: sovling_time}
\renewcommand{\arraystretch}{1.4}
\setlength{\tabcolsep}{0.7mm}{
\footnotesize{
\begin{tabular}{|c|c|c|c|c|c|c|c|c|c|}
\hline
     & \multicolumn{3}{c|}{$|\widetilde{\mathbb{T}}| = 4$}              & \multicolumn{3}{c|}{$|\widetilde{\mathbb{T}}| = 6$}              & \multicolumn{3}{c|}{$|\widetilde{\mathbb{T}}| = 8$}              \\ \hline
$N$  & $\overline{t}^{\rm prune}$ (sec) & $t^{\rm adj}$ (sec) & $t^{\rm ip}$ (sec)      & $\overline{t}^{\rm prune}$ (sec) & $t^{\rm adj}$ (sec) & $t^{\rm ip}$ (sec)       & $\overline{t}^{\rm prune}$ (sec) & $t^{\rm adj}$ (sec) & $t^{\rm ip}$ (sec)       \\ \hline
$5$  & $8.14\pm3.93$              & $0.07\pm0.04$ & $2.05\pm1.53$       & $30.49\pm5.47$             & $0.17\pm0.12$ & $3.78\pm1.89$       & $48.59\pm14.38$            & $0.28\pm0.16$ & $6.43\pm4.20$        \\ \hline
$10$ & $13.7\pm3.12$              & $0.27\pm0.08$ & $7.21\pm3.37$       & $22.16\pm7.16$             & $0.26\pm0.16$ & $8.31\pm6.87$       & $38.53\pm13.04$            & $0.36\pm0.19$ & $8.24\pm5.33$       \\ \hline
$15$ & $12.87\pm2.74$             & $0.35\pm0.15$ & $181.18\pm229.33$   & $25.74\pm5.74$             & $0.56\pm0.28$ & $91.24\pm100.05$    & $38.43\pm13.08$            & $0.63\pm0.26$ & $77.86\pm80.47$     \\ \hline
$20$ & $10.90\pm2.99$              & $0.41\pm0.21$ & $146.72\pm148.56$   & $25.00\pm7.36$              & $0.66\pm0.37$ & $261.13\pm358.53$   & $29.25\pm12.54$            & $0.45\pm0.29$ & $516.50\pm1186.96$   \\ \hline
$25$ & $8.32\pm3.65$              & $0.33\pm0.18$ & $1741.57\pm4160.64$ & $22.79\pm9.21$             & $0.6\pm0.31$  & $808.92\pm1454.72$  & $28.71\pm15.82$            & $0.86\pm0.51$ & $2816.32\pm2164.77$ \\ \hline
$30$ & $5.17\pm2.05$              & \bm{$0.38\pm0.21$} & \bm{$4129.2\pm6706.97$}  & $17.31\pm6.97$             & \bm{$0.6\pm0.28$}  & \bm{$4941.11\pm5694.07$} & $30.05\pm10.86$            & \bm{$0.87\pm0.42$} & \bm{$4271.79\pm8339.85$} \\ \hline
\end{tabular}
}}
\end{table*}

\begin{figure}[t]
    \centering
    \includegraphics[width=\linewidth]{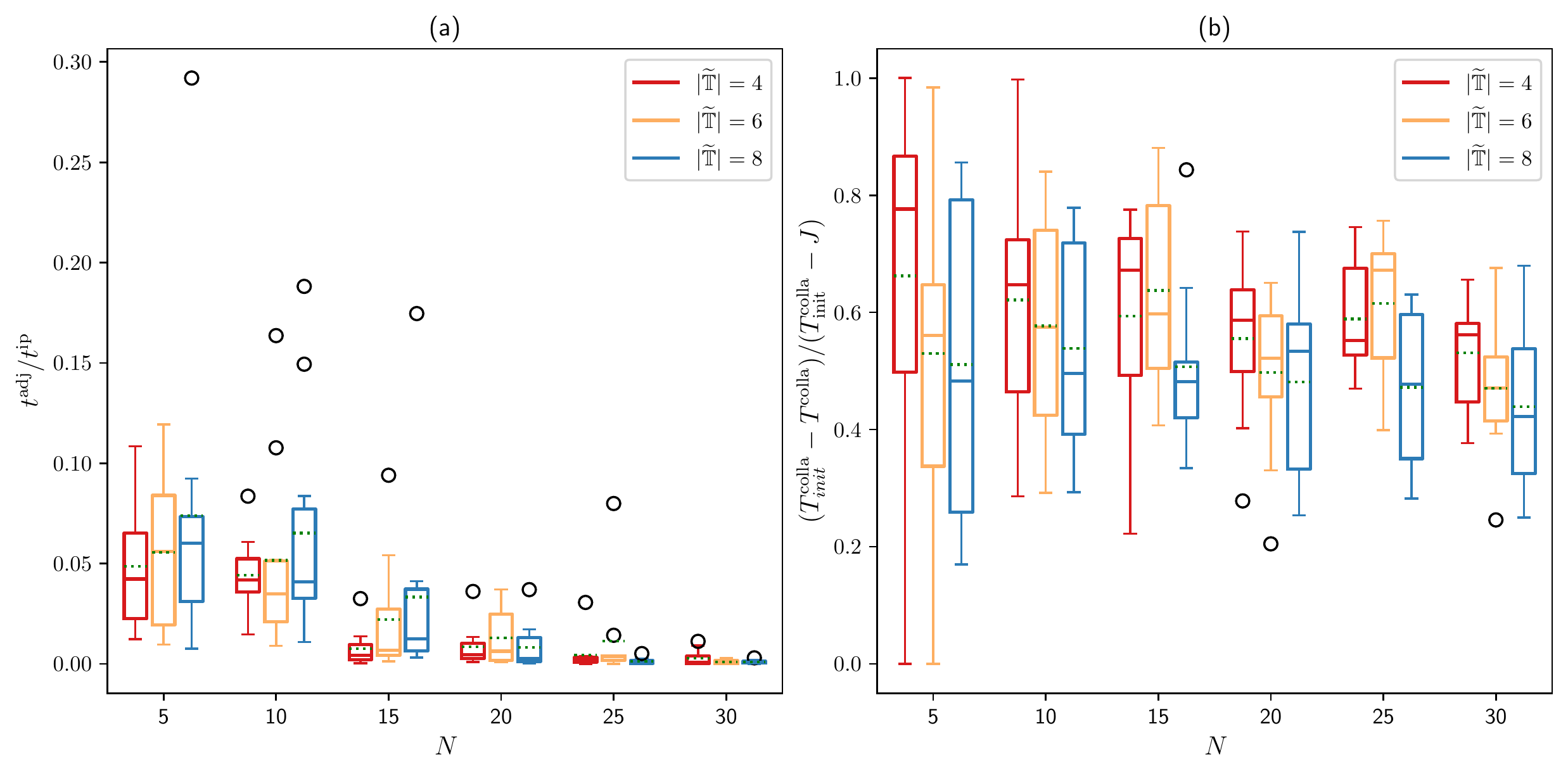}
    \caption{The comparison of the adjusting method and the MILP method under different number of collaborative tasks: (a) $t^{\rm adj}/t^{\rm ip}$; (b) $(T^{\rm colla}_{\rm init} - T^{\rm colla})/(T^{\rm colla}_{\rm init} - J)$. Here $t^{\rm adj}$ denotes the time token by the execution strategy adjusting mechanism, $t^{\rm ip}$ denotes the time token by Gurobi to solve the MILP problem. Each entry shows the average (green dotted line), the median, 25th percentile, 75th percentile, the max, the minimum and outliers out of 10 experiments.}
    \label{fig:adj_vs_ip_ct}
\end{figure}

We conduct extensive simulations to investigate the performance of the execution strategy adjusting mechanism versus the baseline MILP method.
The number of robots varies from $\{5, 10, 15, 20,25,30\}$ and the number of collaborative tasks changes from $4$, $6$ to $8$.
We also investigate the performance under three types of environment sizes.
In each case, we run both the proposed framework with adjusting mechanism for $10$ times, where all tasks are randomly distributed in the grid map environment for $10$ times under the same task specifications. 
For the proposed framework with adjusting mechanism, we stop the simulation once the running time exceeds $30$ minutes, and select the best result obtained up to the end time. 
We then compare the running time and the solution quality of the proposed framework with adjusting mechanism versus with MILP method under the same assignment of collaborative tasks.

In Fig.~\ref{fig:adj_vs_ip_ct}$(a)$, we compare the solving time of the proposed adjusting mechanism with the MILP method under different number of robots and different types of collaborative tasks.
It is obvious that in all cases, the adjusting mechanism runs much faster than the MILP method, and the advantage of the adjusting mechanism becomes significant as robot number increases.
Surprisingly, the adjusting mechanism stops within one second in all cases, while the MILP method takes more than $10$ hours to solve the case with $30$ robots.
The detailed comparison is shown in Tab.~\ref{table: sovling_time}.
Furthermore, in Fig.~\ref{fig:adj_vs_ip_ct}$(b)$, we evaluate how much the total time cost can be optimized by the proposed adjusting mechanism. This is evaluated by $(T^{\rm colla}_{\rm init} - T^{\rm colla})/(T^{\rm colla}_{\rm init} - J)$, where $T^{\rm colla}_{\rm init}$ corresponds to the total time cost of robots' initial execution strategies before adjustment, $T^{\rm colla}$ is the total time cost after applying the adjusting mechanism and $J$ is the result of the MILP method.
As in Fig.~\ref{fig:adj_vs_ip_ct}$(b)$, the optimization effect of the adjusting mechanism varies when the problem is small, however, it tends to converge within around $40\%$ to $60\%$ as the number of robots increases, which verifies that the effect of proposed adjusting mechanism to reduce the total time cost.

Specifically, the detailed time spent for solving each case are shown in Tab.~\ref{table: sovling_time}. 
The table compares the average pruning time ($\overline{t}^{\rm prune}$) for robots to prune their local product automaton as in Def.~\ref{def: prune_local_product}, the time spent by the adjusting mechanism ($t^{\rm adj}$) and the time spent by the MILP method ($t^{\rm ip}$) under different numbers of collaborative tasks and numbers of robots.
Note that pruning robots' local product automatons is the common preliminary step before applying the adjusting mechanism or the MILP method.
The average time spent by robots to prune their local product automatons, $\overline{t}^{\rm prune}$, grows slightly as the number of collaborative tasks increases from $4$ to $8$.
This is because on average, the robots are assigned more collaborative tasks, so that the size of their pruned product automatons also creases as in Prop.~\ref{prop_prunecomplexity}.
The same trend also exists in $t^{\rm adj}$ and $t^{\rm ip}$, despite some exceptions when the robot population is small.

Thanks to the preliminary procedure of pruning local product automatons, the MILP method can get the results within several seconds when the robot number is small.
However, $t^{\rm ip}$ grows significantly as the number of robots increases.
Note that in a case with $N=30$ and $|\tildeT| = 8$, the MILP method takes more than $7$ hours to get the solution, while the adjusting mechanism only takes $1.93$ seconds to converge.
Surprisingly, the proposed execution strategy adjusting mechanism solving almost all cases within a second (maximum value is $2.24s$).

\begin{figure}[t]
    \centering
    \includegraphics[width=\linewidth]{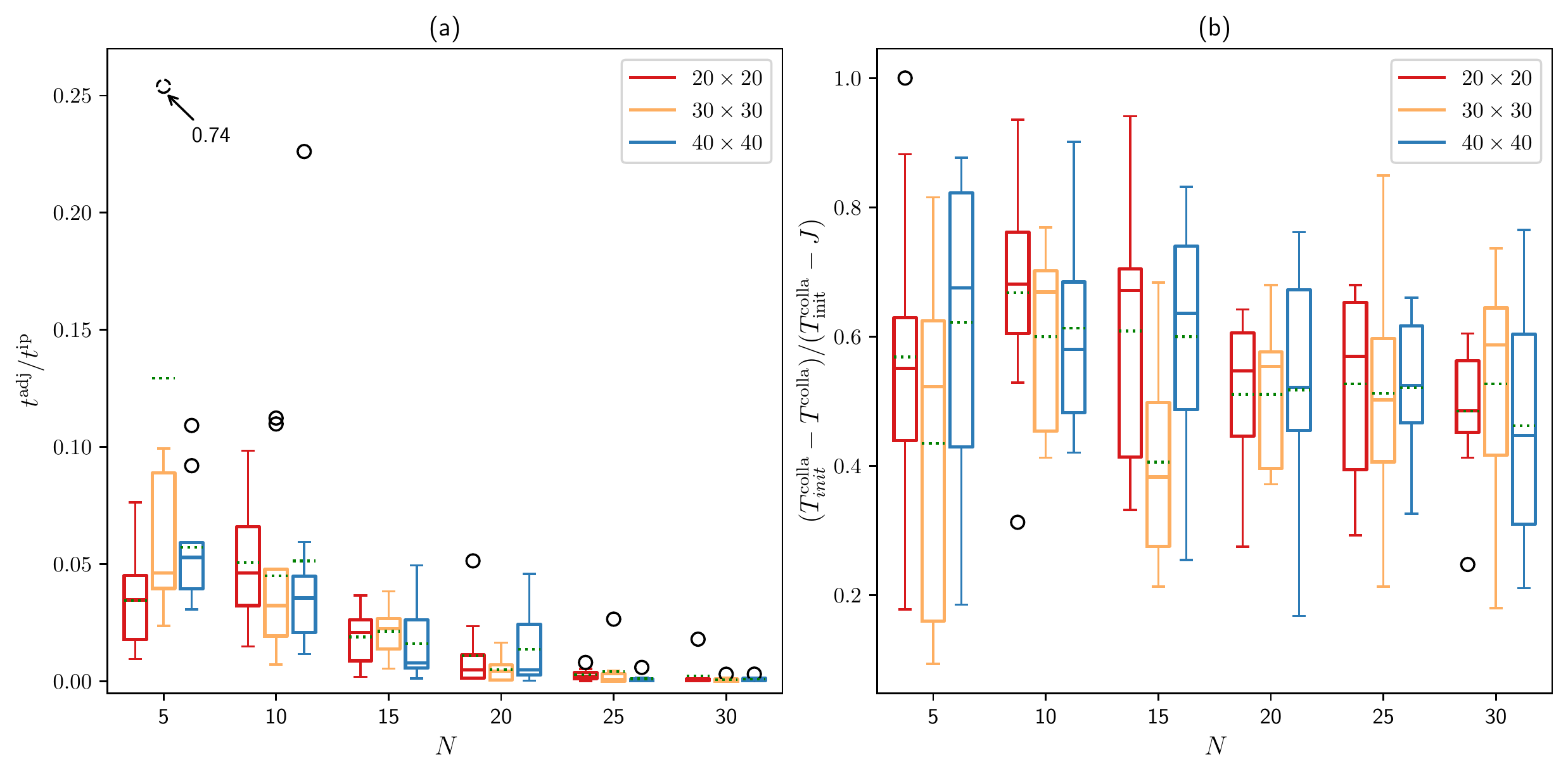}
    \caption{The comparison of the adjusting method and the MILP method under different environment sizes: (a) $t^{\rm adj}/t^{\rm ip}$; (b) $(T^{\rm colla}_{\rm init} - T^{\rm colla})/(T^{\rm colla}_{\rm init} - J)$.
    Each entry shows the average (green dotted line), the median, 25th percentile, 75th percentile, the max, the minimum and outliers out of 10 experiments.}
    \label{fig:adj_vs_ip_env}
\end{figure}

Moreover, we also compare the above metrics of the two methods under different environment sizes, varying from $20\times 20$, $30\times 30$ to $40\times 40$ grid map. The results are shown in Fig.~\ref{fig:adj_vs_ip_env}.
The performance of the proposed method is consistent with it in Fig.~\ref{fig:adj_vs_ip_ct}. 
The performance of the proposed methods shows no obvious difference under different environment size in Fig.~\ref{fig:adj_vs_ip_env}, because the environment size doesn't influence the sizes of the pruned local product automatons according to Prop.~\ref{prop_prunecomplexity}.
The environment size may change the time robot spend to move between different regions, and we can see from Fig.~\ref{fig:adj_vs_ip_env}(b) that such change doesn't obviously affect the optimization effect.

\subsection{Iterative Optimization and Filtering of SMT Solutions}

\begin{figure}[t]
    \centering
    \includegraphics[width=\linewidth]{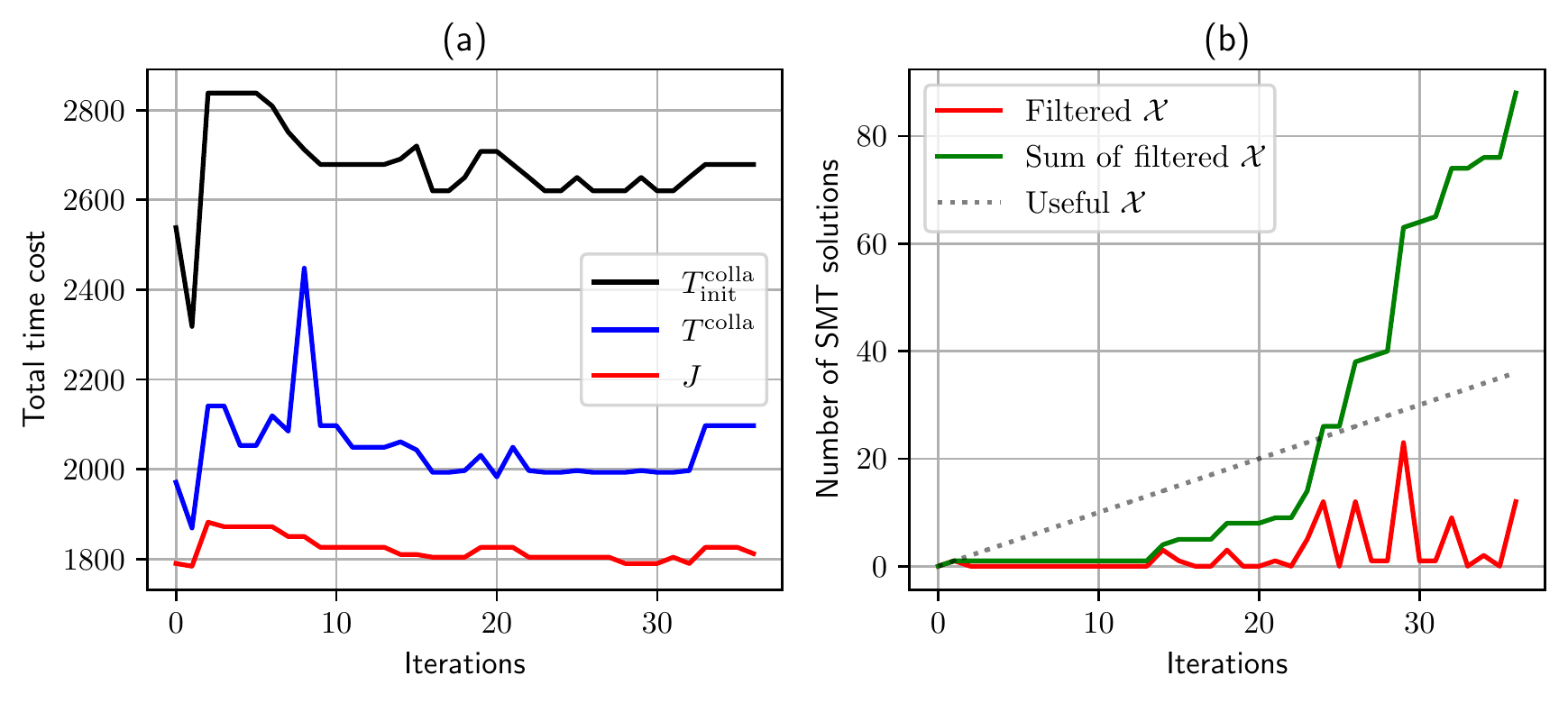}
    \caption{(a) The optimization process of the total time cost under different assignment of collaborative tasks and (b) the filtered SMT solutions in each iteration by utilizing the filtering strategy in an experiment with $15$ robots.}
    \label{fig:one_case}
\end{figure}

The proposed framework iteratively evaluates each feasible assignments of collaborative tasks, which is generated by solving an SMT model (Sec.~\ref{subsec: smt}). And we also propose a filtering strategy to screen out non-optimal assignments in advance.
Fig.~\ref{fig:one_case} $(a)$ plots the optimization process of the total time cost under each possible assignment of collaborative tasks generated by the SMT solver, in a case with $15$ robots, $|\tildeT|=6$ and environment size is $30\times 30$.
In each iteration (each kinds of feasible task assignments), the total time cost is reduced from $T^{\rm colla}_{\rm init}$ to $T^{\rm colla}$ by the proposed adjusting mechanism.
Although the MILP method can provide better solutions in each iteration, as shown by the red line, it takes much more time than the adjusting mechanism to get the results.
In addition, as shown in Fig.~\ref{fig:one_case} $(b)$, the filtering strategy proposed in in Sec.~\ref{subsec: smt} can efficiently filter non-optimal solutions to the collaborative task assignment problem according to the history solutions, avoiding unnecessary computation.
The strategy only needs to maintain a set of solutions whose size is propositional to the number of iterations.
Note that the filtering efficiency depends on the inner computation mechanism of the SMT solver and the specific task requirements. 

\section{Conclusions and Future Work}
\label{sec: conclusion}
In this paper, we propose a hierarchical task planning framework that can efficiently coordinate multiple robots under individual and collaborative temporal logic specifications.
A central server first extracts a task sequence satisfying the collaborative task specification, decomposes it, and then allocates subtasks to the robots. 
All feasible task assignments are iteratively generated by an SMT solver and evaluated by the robots.
Then the robots synthesize their execution strategies based on locally constructed product automatons. 
To further optimize the total time cost, we propose an distributed execution strategy adjusting mechanism that allows robots adjust their strategies to minimize inefficient wait time in collaborations.
We prove the completeness of the proposed method under our assumptions and extensive simulation results verify the scalability and efficiency of the proposed method.
Future work is to improve the efficiency of the task allocation procedure and expand the current framework to complete LTL to enable infinite tasks.

\addtolength{\textheight}{-12cm}

 \bibliographystyle{elsarticle-num} 
 \bibliography{root}

\end{document}